\documentclass[journal,twoside]{IEEEtran}

\normalsize

\usepackage[cmex10]{amsmath}
\usepackage{graphicx}
\usepackage{epstopdf}
\usepackage{multirow}
\usepackage{amssymb}
\usepackage{ifthen,boxedminipage}
\usepackage{cite}
\usepackage{lipsum}
\usepackage{mathtools}
\usepackage{cuted}
\usepackage{varwidth}
\usepackage{tabularx}
\usepackage{rotating}
\usepackage{times,color}
\definecolor{dgreen}{rgb}{0,0.48,0.3}

\begin{document}

\title{Enhancing ensemble learning and transfer learning in multimodal data analysis by adaptive dimensionality reduction}

\author{Andrea Marinoni, Saloua Chlaily, Eduard Khachatrian, Torbjørn Eltoft, Sivasakthy Selvakumaran, Mark Girolami, Christian Jutten 
	\thanks{A. Marinoni is with Dept. of Physics and Technology, UiT the Arctic University of Norway, P.O. box 6050 Langnes, NO-9037, Tromsø, Norway, and Dept. of Engineering, University of Cambridge, Trumpington St., Cambridge CB2 1PZ, UK. E-mail: andrea.marinoni@uit.no. 
	S. Chlaily, E. Khachatrian, and T. Eltoft are with Dept. of Physics and Technology, UiT the Arctic University of Norway, P.O. box 6050 Langnes, NO-9037, Tromsø, Norway. E-mail: $\{$saloua.chlaily, eduard.khachatrian, torbjorn.eltoft$\}$@uit.no. 
	S. Selvakumaran is with Dept. of Engineering, University of Cambridge, Trumpington St., Cambridge CB2 1PZ, UK. E-mail: ss683@cam.ac.uk. 
	%T. Damoulas is with Dept. of Computer Science and Dept. of Statistics, University of Warwick, Coventry CV4 7EZ, UK, and with The Alan Turing Institute, 96 Euston Rd, London NW1 2DB, UK. E-mail: tdamoulas@turing.ac.uk. 
	M. Girolami is with Dept. of Engineering, University of Cambridge, Trumpington St., Cambridge CB2 1PZ, UK, and with The Alan Turing Institute, 96 Euston Rd, London NW1 2DB, UK. E-mail: mag92@cam.ac.uk. 
	C. Jutten is with GIPSA-lab, University of Grenoble Alpes, 38000 Grenoble, France, and with the Institut Universitaire de France, 75005 Paris, France. E-mail: christian.jutten@gipsa-lab.grenoble-inp.fr.	
	}
}

\maketitle

\begin{abstract}
Modern data analytics take advantage of ensemble learning and transfer learning approaches to tackle some of the most relevant issues in data analysis, such as lack of labeled data to use to train the analysis models, sparsity of the information, and unbalanced distributions of the records. 
Nonetheless, when applied to multimodal datasets (i.e., datasets acquired by means of multiple sensing techniques or strategies), the state-of-the-art methods for ensemble learning and transfer learning might show some limitations. 
In fact, in multimodal data analysis, 	not all observations would show the same level of reliability or information quality, nor an homogeneous distribution of errors and uncertainties. 
This condition might undermine the classic assumptions ensemble learning and transfer learning methods rely on. 
In this work, we propose  an adaptive approach for dimensionality reduction to overcome this issue. 
By means of a graph theory-based approach, the most relevant features across variable size subsets of the considered datasets are identified. This information is then used to set-up ensemble learning and transfer learning architectures. 
We test our approach on multimodal datasets acquired in diverse research fields (remote sensing, brain-computer interfaces, photovoltaic energy). 
Experimental results show the validity and the robustness of our approach, able to outperform state-of-the-art techniques. 
\end{abstract}

%\begin{keywords}
%Multimodal data analysis, ensemble learning, transfer learning, dimensionality reduction, adaptive data analysis.	
%\end{keywords}

\section{Introduction} 

The recent technological advancements in sensing and measuring have induced a data deluge in several domains, from physics to biology, to chemistry and Earth science \cite{multimod1,capacity,multimod2,multimod3,multimod4,multimod5,multimod8}. 
This means that natural and anthropogenic phenomena can be described by several datasets collected in different domains. 
Consequently, the development of efficient, robust, and accurate methods for information extraction has become a hot topic in data analysis as private and public groups working in environmental, biomedical, welfare, energy, and military sectors can take advantage of the deep analysis of the multivariate and multirelational records collected by multiple sensing devices \cite{DDFE_1,multimod1,multimod6,multimod7}. 

This interest is the result of an intrinsic limitation of measuring tools, for which it is rare that a single acquisition method provides
complete understanding about a phenomenon or a system of interest \cite{multimod5,multimod8,multimod3}. 
Indeed, it is true that natural and anthropogenic  processes and systems can be very complex, so that different types of instruments, measurement
techniques, experimental setups, and other types of sources of information might be necessary to provide a complete characterization of their relevant properties and conditions.  
Thus, investigating multiple datasets  collected from different acquisition methods pertaining to a given system or phenomena could extract a tremendous amount of information more than is achievable by analyzing each dataset separately \cite{multimod1,capacity}. 

Every dataset is the product of an acquisition framework (\textit{modality}), that is characterized by a unique set of sensing and measurement techniques.  
Hence, exploring the records collected by each modality implies investigating \textit{multimodal} systems \cite{multimod1}. 
As each modality works under specific physical principles and domains, it is possible to state that a key factor in multimodal data analysis is \textit{diversity}, since each modality brings to
the whole some type of added value that cannot be deduced
or obtained from any of the other modalities in the setup. 
As a consequence, addressing the complementarity provided by the diverse modalities in the data investigation enables the analysis to improve the characterization of the underlying phenomena by providing a global overview of their complex properties.
In other terms, diversity enables the achievement of deeper interpretability, accuracy, robustness, and stability in data analysis by inducing constraints that could be used in defining the feasibility set of the multivariate solutions. 
It therefore identifies the optimal solutions of the analysis problems in a more precise manner  \cite{multimod1,multimod9,multimod6,machlearPR1,mackaybook,manifoldPR1}. 

For these reasons, multimodal data analysis has become a crucial factor in a number of physical, biological, environmental, and eventually sociological application scenarios\cite{multimod3,multimod4,multimod5,multimod8,multimod7,DDFE_1}. 
In fact, multimodal data analysis can help in extracting relevant information from the data for multiple purposes and end-users, so that knowledge can be built and decision can be safely taken. 
This ability can lead to enhance the exploration of the records for the identification of significant parameters or precursors of a given phenomenon, as well as detecting anomalies in the system and measuring the impact of an event in the dynamic of the considered scenario \cite{multimod1,multimod2}. 
All these characteristics make multimodal data analysis the benchmark for multidisciplinary research fields, that are nowadays encompassed by the framework of the Sustainable Development Goals defined by United Nations. 
In this context, the Green Deal policies promoted by the leading countries and institutions worldwide represent a concrete example of how society at large and the natural environment can benefit by the solid understanding and assessment of the human-environment interaction delivered by thorough multimodal data analysis \cite{greendeal,capacity,greendeal2}. 

It is also true that multimodal data analysis represents one of the most challenging sectors in the information technology community \cite{multimod1,capacity,multimod9,multimod2}. 
In fact, data generated by very complex
systems, driven by numerous underlying processes
and that depend on a large number of variables are difficult to directly evaluate and compare. 
%Moreover, investigating records collected by heterogeneous sources of information implies that we need to compare these quantities, which is not a trivial task. 
It is worth noting that large scale heterogeneous datasets are typically considered when performing multimodal data analysis in order to fully exploit its ability to characterize complex phenomena, hence enabling global scale understanding of the considered problem \cite{multimod3,multimod4,multimod5,multimod6,multimod7}. 

The aforesaid challenges translate into various issues from a data analysis perspective \cite{multimod1}, e.g.:
\begin{itemize}
	\item different types of acquisition methods and observation setups provide data at different sampling points, and often at very disparate resolutions; 
	\item measurements may be represented by different types of physical units that are not directly comparable (e.g., reflectance [energy] vs. misplacement [meter]);
	\item different acquisition techniques and experimental setups lead to different amount of samples provided by each modality;
	\item scarcity of high quality samples that can be used to learn underlying characteristics of the considered phenomena across the modalities; 
	\item other complicating factors include:
	different types of noise and spatial distortions;
	varying contrasts; 
	different positions of the observing/measuring instruments.
\end{itemize}
Consequently, these issues lead to investigating datasets that show different statistical and semantic distributions in terms of noise and informativity balance \cite{multimod1}. 
Therefore, employing an approach that is able to address these points is a key component for the extraction of significant information from multimodal datasets. 
In fact, traditional machine learning methods may
fail to obtain satisfactory performances when dealing with
complex data, such as imbalanced, high-dimensional, noisy
data, despite the great achievements obtained in the last decade in this research field \cite{multimod2,multimod9,capacity}.

Transfer learning and ensemble learning are the new trends for solving the problems that arise from training data and test data having different distributions \cite{RFsurvey1,SVMpattern2,multimod2,multimod9,ELsurvey1}. 
It has been proven that these approaches are able  to capture multiple characteristics of  underlying
structure of data. 
These classes of algorithms are able 
 to effectively construct
an efficient knowledge discovery and mining model in several data investigation cases. 
Moreover, by taking advantage of linear programming suites, these methods show high efficiency, one of the key reasons for their success in the data science community.   

Although these algorithms are intrinsically versatile and flexible, they might show some limitations when applied to multimodal data analysis \cite{multimod1,multimod2,multimod9}. 
In fact, when multiple modalities are explored, each acquisition method produces heterogeneous types of errors and uncertainties. 
Thus, 
not all observations or data entries have the same level of confidence, reliability or information quality \cite{mackaybook}. 
This condition might undermine the classic data mining scheme ensemble learning and transfer learning rely on. 
Specifically, in these data analysis schemes the reliability of the data is addressed only as statistical relevance with respect to the learning procedure (for instance, by majority voting the decision to use to label the samples, and by transferring information between samples on the minimum distance path) \cite{multimod2,capacity,mackaybook}. 

In this paper, we aim to enhance the classic ensemble learning and transfer learning workflow implemented for multimodal data analysis by means of an adaptive dimensionality reduction scheme based on a double graph Laplacian representation of the multimodal datasets. 
By selecting the most relevant attributes across the modalities for each sample, it is possible to take advantage of the diversity of the datasets to enhance the accuracy of data analysis, while 
tackling the heterogeneity of the noise and uncertainties within the large-scale data clouds. 
In this way it is possible to 
overcome issues associated with a lack of training data, so to fully exploit the scarce subsets of labeled records for training and setting the learning models.

This paper is organized as follows. 
Section \ref{sec_methback} provides a thorough introduction of the related works and the main motivations and contributions of this work.   
Section \ref{secmeth} introduces the proposed method and its implementation details.  
Section \ref{secresult} reports the performance results. 
Finally, Section \ref{secconcl} delivers our final remarks and some ideas on future research. 

\section{Background $\&$ motivation}
\label{sec_methback}

As previously mentioned, ensemble learning and transfer learning are research topics that have been widely studied by the scientific community in the last decades. 
In this Section, we provide a brief summary of the main approaches  that have been introduced in technical literature. 
Specifically, we summarize the main aspects of schemes that aim to extract information by ensemble methods employed at model-level (i.e., multiple learners used to improve the characterization of the data) and at data-level (i.e., algorithms exploiting diversity by data partitioning), as well as the main properties of the major approaches for transductive transfer learning.  
We focus our attention especially on structures that take advantage of dimensionality reduction schemes to improve efficiency and accuracy. 
Finally, we introduce the main motivations that drive this work, and its main contributions to the scientific community. 

\subsection{Main approaches in model-level ensemble learning}

Ensemble learning encompasses the use of 
 a collection of several classifiers whose individual
decisions are combined to classify the test examples \cite{RFsurvey1,RFsurvey2,RFsurvey3,RFsurvey4,RFsurvey5,ELsurvey1}. 
It is known that an ensemble
often shows a much better performance than the individual classifiers that compose
it. 
%Indeed, it has been proven that diversity can enhance classification results . 
%Specifically, let us assume that 
%$N_C$ different classifiers are used to investigate the characteristics of a dataset: let us also consider that their
%errors are uncorrelated, and their individual probability of error is less than 1/2. 
%In this case, the classification achieved by  the combination of their outcomes (also in terms of majority voting) decreases as $N_C$ gets large 
\cite{RFsurvey2,RFsurvey5}.
% 
%Hansen and Salamon [13] shows why the ensemble shows better performance
%than individual classifiers as follows. 
%Assume that there are an ensemble of N classifiers:
%f f1; f2; . . . ; fNg and consider a test data x. If all the classifiers are identical,
%their results are matched on the same data, therefore an ensemble will show the same
%performance as individual classifiers. 
%However, if classifiers are different and their
%errors are uncorrelated, then when fi is wrong, most of the other classifiers except
%for fi may be correct. If so, the result of majority voting can be correct. More
%precisely, if the error of individual classifier is p < 1=2 and the errors are independent,
%then the probability pE that the result of majority voting is incorrect
%is . When the size of
%classifiers N is large, the probability pE becomes very small. 
%
%Grouping identical learners can bring no gain in classification capacity, and the predictive
%ability can be improved by integrating those classifiers who make errors on different samples. 
%
%Therefore, integrating the properties of diverse classifiers to analyze a dataset can improve the classification performance. 
%Thus, when considering ensemble learning, b
Bagging is the main approach employed in ensemble learning because of its very simple implementation \cite{RFsurvey3,RFsurvey1,RFsurvey2}. 
Bagging consists of two phases:
\begin{itemize}
	\item bootstrap: many data subsets are obtained by sampling with
	replacement on the training data set;
	\item weak
	classifiers (i.e., classifiers based on very simple discrimination criteria, e.g., thresholds or linear hyperplanes) learned on subsets are aggregated by majority
	voting.
\end{itemize}

\noindent %Several instances of bagging-based algorithms for ensemble learning have been proposed in technical literature. 
Although its many advantages, bagging is indeed very sensitive to training data, which might limit its effectiveness (see for instance \cite{randfor6,randfor7,randforPR5}). 

Another widely used ensemble learning method is boosting \cite{RFsurvey1,RFsurvey2,RFsurvey5}. 
It involves two main steps:
\begin{itemize}
	\item many classifiers are learned on training data
	by a given weak learner (i.e., a simple model that do only slightly better than random chance); 
	\item the classifiers produced
	by the weak learner are combined into a single composite classifier.
\end{itemize}
\nonumber In this way, boosting is able to enhance the performance of classification, while ensuring higher robustness to the training phase with respect to bagging.  
%Among examples of boosting schemes, 
AdaBoost is the most famous example \cite{RFsurvey2,RFsurvey5,RFsurvey1}. 
It employs self-adaptation to training data, and adjusts the weights of the classification process so that more attention is put on the samples showing high variance at each iteration. 
Moreover, the combination of an arbitrary number of classifiers performed at each iteration makes AdaBoost typically robust to overfitting (even after a large number of iterations), as well as capable of reducing  the generalization error. 
For these reasons, AdaBoost has been successfully employed in several research fields, from face recognition to biomedical imaging, from natural language processing to remote sensing data analysis \cite{RFsurvey2,RFsurvey5,randforPR4}. 

These approaches have been proven to be effective and simple to implement. 
In fact, several classifiers based on diverse data analysis principles (from decision trees to support vector machines) can be used within these frameworks. 
Ensemble learning implemented by exploiting the diversification of classifiers can help in reducing the information extraction issues each single classifier might have (e.g., it would reduce the effect of overfitting when decision trees are employed). 
However, it is still hard to define a clear relationship between diversity and accuracy that can be achieved \cite{RFsurvey2,RFsurvey5,ELsurvey1}. 
%In fact, increasing diversity among learners
%leads the loss of the average precision of all the classifiers \cite{multimod1}. 
Specifically, the aforementioned methods imply that 
every
subspace contains enough informative features for training
better classifiers and increasing diversities of classifiers.
This is
crucial to improve the performance of an ensemble classification model effectively in terms of prediction accuracy
%This is crucial to effectively improve the performance
%of an ensemble classification model in terms of prediction
%accuracy 
\cite{ELsurvey1,RFsurvey2}.

It is important to find an adequate balance
between diversity and precision of these elements, in order for the ensemble scheme to robustly gain 
precision. 
This problem can be thoroughly addressed only in combinatorics terms, which can then hardly be investigated by exhaustive search. 
Pruning identifies a viable solution to this point \cite{ELsurvey1}. 
Given a set of trained individual learners, rather than combining all of
them, ensemble pruning tries to select a subset of individual learners to
comprise the ensemble, so to enhance the extraction of relevant information while avoiding the precision of the ensemble scheme to degrade. 
Moreover, pruning is a key factor in enhancing the efficiency of ensemble learning schemes, especially when classifiers showing high order of generalization (e.g., support vector machines - SVMs) are employed. 
%In fact, this characteristic implies strong effort in terms of computational cost. 
On the other hand, pruning might result in a strong computational cost and a limiting factor for the use of such architectures, especially in operational scenarios where near real-time operations are requested \cite{ELsurvey1,RFsurvey2,RFsurvey5,multimod7}.  
%
% To find the approximation solution, many pruning methods were
%proposed in literature (see the Sect. 1.1).
%According to the analysis, it shows the pruning methods utilizing diversity measures, combined
%with different pruning ways, cannot find the exact solution

Pruning methods can be categorized into three groups: ordering-based pruning, optimization-based pruning, clustering-based pruning. 

\subsubsection{Ordering-based pruning}
These schemes aim to assess the relevance of each weak classifier by using a metric to sort the significance of their outcomes \cite{ELsurvey1,ELmodel_pruning1,ELmodel_pruning2,ELmodel_pruning3}. 
This approach implies the definition of a metric to evaluate the outcomes, and consequently the definition of a criterion for thresholding and aggregation. 
Methods in the technical literature use metrics such as reduce-error distance, complementarity measure,
margin distance, Kappa statistics \cite{ELsurvey1,ELmodel_pruning1,ELmodel_pruning2,ELmodel_pruning3}. 
Moreover, the selection of weak classifiers can be achieved by employing diverse strategies to rank their actual contribution to the analysis (e.g., in terms of a hard thresholding based on margin-based criterion, or in terms of a diversity-based distance) in order to achieve the final set of the weak classifiers that can be considered to generate the final output of the ensemble scheme \cite{ELsurvey1}. 
%
%The different pruning methods employ different measures for ordering. Martínez-Muñoz
%et al. [9, 27, 33] used different measures such as reduce-error, complementarity measure,
%and MDM to gain the selection of learners. Experiment results express that they can obtain
%higher accuracies than Bagging. Margineantu et al. [34] selected the learners with theirKappa
%measures less than a certain value for aggregation. Lu et al. [35] proposed an effective pruning
%method, the selection of learners is achieved using the ranks in terms of their contribution.
%Guo et al. [36] proposed a pruning technique, namedMOAG, it extracts the learners with their
%margin-based criterion over a predefined value. Guo et al. [37] put forward a new pruning
%method named MDEP, and it uses the margin and diversity-based measures of learners to
%gain the final ensemble.

\subsubsection{Optimization-based pruning}
These methods describe the pruning problem in terms of a combinatorial optimization problem \cite{ELmodel_pruning4,ELsurvey1}. 
In this respect, methods based on a genetic algorithm approach have been widely used within the scientific community: the weak classifiers are selected according to the weights that are used by the genetic optimization trajectory to find the best subset of classifiers within the total ensemble\cite{ELmodel_pruning4,ELmodel_pruning5}. 
Methods based on glowworm swarm optimization are part of this category too, as well as methods that use collective
agreement to perform pruning \cite{ELmodel_pruning6,ELmodel_pruning7}. 
In the latter case, each classifier is assessed  by calculating the first agreement measure
between its predictive results and the actual labels, and the second measures between any
two learners are also obtained. 
According to these quantities the classifiers are selected to be part of the ensemble learning process. 

\subsubsection{Clustering-based pruning}
The methods that are grouped in this category seek for representative classifiers by segmenting the learners in sub-ensembles \cite{ELsurvey1,ELmodel_pruning8,ELmodel_pruning9}. The main steps of clustering-based pruning work under the principal assumption for which the classifiers can be grouped in diverse clusters. Then, high diversity in the ensemble learning structure can be achieved by selecting learners from various clusters. To this aim, the classifiers close to the cluster centroids should be used to build the pruned ensemble.  
Within this category, 
several methods have been used to perform pruning, such as k-means \cite{ELmodel_pruning8}, deterministic
annealing \cite{ELmodel_pruning9}, hierarchical clustering \cite{ELmodel_pruning10}. 

\subsection{Main approaches in data-level ensemble learning}

Exploiting diversity to investigate multimodal datasets might be particularly cumbersome when imbalanced datasets are analyzed \cite{ELdata_pruning5,ELdata_pruning1}. 
In this case, the final decision of the classification scheme might be biased by the non-uniform distribution of the samples, either in terms of statistical characteristics (e.g., multiple noise sources  and/or inconsistent data are present in the dataset) or in terms of semantic properties (e.g., some classes occurring more often than others within the considered set). 
In fact, when one or more classes outnumber the others within the dataset, it is very hard for the data analysis framework to properly generalize the actual properties and conditions of the samples and hence hard to improve the understanding of the phenomena described by the collected records \cite{ELdata_pruning5,ELdata_pruning1}. 

Employing multiple classifiers focused on multiple partitions of the original dataset can help in dealing with statistical and semantic imbalance across the samples. 
Thus, multiple sub-datasets are built according to cluster partitioning methods so that the occurrence of all classes is kept as uniform as possible and the samples within each subset would show consistent statistical characteristics. 
In this way, the problem of imbalanced dataset can be reduced. 

Several methods can be utilized to perform data partitioning. 
The main approaches can be summarized as follows \cite{ELdata_pruning1,ELdata_pruning2,ELdata_pruning3,ELdata_pruning4,ELdata_pruning5}:
\begin{itemize}
	\item \textit{sampling methods}: samples are randomly grouped in order to make the semantic distribution of the samples within the subsets as uniform as possible. Typically, a criterion for homogeneity is required to drive the grouping operation;
	\item \textit{oversampling methods}: these methods first find the \textit{k} nearest
	neighbours of each sample under exam, depending on the amount
	of oversampling required to make the samples as uniformly distributed as possible;
	\item \textit{separation methods}: these schemes aim to segment the dataset in coherent subsets according to an adjacency criterion (which could be defined in  graph- or continuous domain-based representations of the datasets). In this way, the samples in each resulting subset show homogeneous characteristics;
	\item \textit{subspace projection methods}: these schemes map the original dataset onto a lower dimensional subspace, where clustering is then performed (e.g., by eigenanalysis or forward-backward selection). 
	
\end{itemize}

Once data partitioning has been applied, each subset is analyzed by means of a properly trained classifier. Support vector machine (SVM) is typically employed in this phase \cite{ELdata_pruning1,ELdata_pruning5,ELdata_pruning6}.
In fact, the ability of SVM to generalize the characterization over large scale datasets by taking into account small scale datasets for training make this approach very appealing,  especially for ensemble learning of partitioned dataset. 
SVM tends to work robustly over small subsets extracted over the full dataset, and can therefore be adequately employed to address diversity by data partitioning.

\subsection{Main approaches in transductive transfer learning}

Transfer learning is a very important task, especially when it is expensive or impossible to
re-collect the required training data and rebuild the models to extract information.
%When the distribution changes, most statistical models need to
%be rebuilt from scratch using newly collected training data. In
%many real world applications, it is expensive or impossible to
%re-collect the needed training data and rebuild the models
%Indeed, several instances of transfer learning can be defined according to the goals and objectives of the considered application. 
%Specifically, 
To consider transfer learning, let us define a source domain as a subset of the considered dataset where full knowledge of the class labels is available. 
Meanwhile we shall define a target domain as the complementary subset of the source domain. 
At this point, we can define transfer learning aims to help improve the learning in the target domain using the knowledge acquired in the source domain \cite{TTL1,multimod1}. 

This task can be implemented in several ways and according to diverse conditions. 
In this work, we are interested in the so-called transductive transfer learning, where the source
and target tasks are the same in terms of information extraction, while the source and target domains are different \cite{TTL1}.
In this situation, no labeled data in the target domain
are available while labeled data in the source
domain are available. 
It is important to note that the marginal distributions of the data in the source and target domains can be different. 
Indeed, the only requirements in transductive transfer learning setting are: $i)$ that part of the unlabeled
target data be seen at training time in order to obtain the
marginal probability for the target data;
 $ii)$ the learning tasks must be the same in terms of information extraction (e.g., classification, target detection, delineation); $iii)$ and there
 must be some unlabeled data available in the target domain, so that one can adapt the predictive function learned
 in the source domain for use in the target domain through
 some unlabeled target-domain data.
 In this respect, transductive transfer learning involves the characteristics of domain adaptation, where the difference
 lies between the marginal probability distributions of source
 and target data; i.e., the tasks are the same but the domains
 are different \cite{TTL1,TTL2}. 
 
Importance sampling has traditionally played a key role in the set-up of the main transductive transfer learning architectures, such as empirical risk
minimization (ERM) \cite{TTL2}. 
In general, the goal is to learn 
the optimal parameters of the model by minimizing the
expected risk, defined according to statistical losses on the observation parameters. 
Moreover, in the transductive transfer learning setting, we want to
learn an optimal model for the target domain by minimizing
the expected risk. 
Kernel mapping is also considered within this approach \cite{TTL1,TTL3}. 
One of the major examples of this strategy is represented by the
kernel-mean matching (KMM) algorithm, which aims to learn the characteristics of the target domain from the source domain 
 directly by matching the means between the source
domain data and the target domain data in a reproducing-kernel
Hilbert space \cite{TTL3}. 
Another technique that relies on importance sampling
is importance
weighted twin Gaussian processes: in this scheme,
the importance weight function is directly learned by density estimation. Thus, this method is based on the relative unconstrained least-squares importance fitting.

In general, the most popular schemes in transductive transfer learning aim to estimate the ratio between the likelihoods
of being a source or target example.
% in order to reduce the effect of the possible imbalance of the distribution 
To this aim, the 
 use of a domain classifier for independent estimation of the likelihoods have been considered \cite{TTL1,TTL4}. 
Another approach that has been proposed is based on the direct approximation of the ratio between the densities with a Kullback-Leibler importance estimation procedure (KLIEP) \cite{TTL4}. 
Several methods to compute the mismatch between the data distributions in the source and target domains have been proposed 
by means of several metrics and approximations, such as: maximum entropy density estimation to infer the resampling weights; addressing predictive inference under covariate shift by weighting the log-likelihood function;
simultaneous optimization of the weights and of the classifier parameters, in order to preserve
the discriminative power of the new decision boundary \cite{TTL1}.

Another successful approach for transductive transfer learning is represented by domain adversarial methods,  mainly because of their ability to alleviate the distribution shift between the
source and the target domains \cite{TTL6}. 
Typically, these methods
assume an identical label space between the two domains. 
This might lead to a strong limitation of the analysis accuracy, especially for operational scenarios, where the target training set may not contain the complete set of classes. 
To overcome incomplete target label space during
training, domain adversarial methods employing unilateral data alignment have been proposed \cite{TTL5}. 
In this way, 
inter-class relationships of the source
domain are exploited, and the data analysis scheme can align unilaterally the target to the source
domain, so that the distance
between the source features and the pre-trained features is minimized. 

It is also worth noting that statistical learning theory has been widely used for enhancing transductive transfer learning performance \cite{TTL7,TTL2}. 
Joint geometrical and statistical alignment techniques have been used to reduce
the distributional and geometrical divergence between
domains simultaneously by exploiting both the
shared and domain specific features. 
To this aim, the source and target data are mapped 
into respective subspaces. 
After the projections, the variance
of target domain data is maximized to preserve the
target domain data properties, and the domain shift is statistically reduced by minimizing the marginal and conditional
distribution divergences between source and target domains \cite{TTL7}. 

In this context, dynamic domain adaptation (DDA) has been proposed to quantitatively assess the relative importance of each
distribution, hence improving the ability of the architectures to track the actual relationship between source and target domains \cite{TTL8}. 
Using this approach, transfer learning based on Grassmann manifold representation for domain adaptation (manifold DDA - MDDA), and a deep learning network to characterize an end-to-end classifier (DDA network - DDAN) have been proposed. 
Moreover, a deep learning network designed to address transductive transfer learning has been proposed to enhance the adaptation from
source to target and explicitly weakens the influence from the irrelevant source classes \cite{TTL9}. 
This deep residual correction network (DRCN) uses partial domain adaptation to properly define the relationship of the statistical distributions describing source and target domains, so to provide a better matching of the feature distributions of
shared classes between source and target, i.e., improving the characterization performance of the data analysis scheme.

\subsection{Motivation and main contributions of this work}

The approaches that have been previously mentioned aim to extract information in a very efficient and versatile way from massive datasets. 
Although their intrinsic flexibility (either in terms of ensemble learning at data- or model-level, as well as of transductive transfer learning) make these strategies very appealing for multimodal data analysis, it is also true that they might find hard to improve understanding of the characteristics underlying the data, especially when the considered datasets are collected in operational and real life scenarios.
In operational scenarios, many of the assumptions made by each of these algorithms are no longer valid, giving rise to some of the potential issues noted in each section above%. 
%In this case, the issues we previously mentioned for the data analysis schemes will apply, making the main assumptions that each of the algorithms alternatively fail
 \cite{multimod1,multimod2,multimod9,lazer,ELdata_pruning5,ELsurvey1,TTL1,capacity}. 

Specifically, when investigating large scale datasets, the geometrical, statistical and physical characteristics of the records may barely described by one single data model \cite{multimod1,capacity,multimod9,multimod2}. This effect is even more evident when multiple sources of information are integrated to build knowledge on the considered dataset. In fact, in this case, the mismatch in terms of granularity and/or resolution of the diverse modalities might lead to signals and noise components with different properties across the records \cite{multimod7,multimod5}. This situation is clearly evident when considering
remote sensing data, such as satellite imagery using different wave spectrum to image the Earth. Since the morphology of the region of interest, as well as the different working conditions of the sensing devices (either in terms of illumination and imaging capacity, or spectral, spatial, and temporal resolution), could result in images where the observations can be linked to the same physical phenomena on Earth surface in a different way depending on their position in the scenes (e.g., in a mountain area or in a flat region) and/or the conditions of the acquisitions (e.g., during sunshine or in the shadow) \cite{multimod1,capacity}. 
From the point of view of data analysis, this situation implies sparse and often inconsistent datasets to be analyzed. 
Further, when considering the semantics of the records, each modality might provide an unbalanced amount of records with respect to the classes or thematic clusters that are the objective of the analysis \cite{multimod1,ELdata_pruning5,multimod2}.

Investigating datasets showing the aforesaid properties is certainly cumbersome. 
This means that the design of learning architectures that are used to explore such datasets must face a dramatic trade-off between automation, accuracy, and computational complexity \cite{multimod2}. 
In particular, the methods that have been presented in the first part of this Section might fail or produce inaccurate results when exploring sets of records with these characteristics. 
In fact, such algorithms implicitly rely on assumptions in terms of continuity - and in several cases linearity - of the data \cite{multimod2,multimod9}. 
This choice is often motivated by the efficiency  improvement that can be obtained in this way, since metrics based on Euclidean geometry (often linear) are used to carry out the investigation of the data. 
In this case, the computational costs of the data analysis system - also fully automatic - would be strongly reduced, because these operations can be transformed in matricial form, which translates in high computational efficiency. 
However, the accuracy of the analysis framework can be strongly degraded, since these assumptions 
might be hardly matched when multimodal observations are taken into account, as a result of the data sparsity and inconsistency \cite{multimod1,ELdata_pruning5}. 

In order to improve precision without reducing the efficiency of the analysis, the traditional methods in ensemble learning and transfer learning 
rely on the availability of large quantities of \textit{a priori} information that can be used to train the learning models \cite{multimod1,capacity}. 
In this way, the data analysis schemes can achieve high precision, while reducing the computational complexity for the validation step. 
However, it is also true that achieving reliable understanding on large subsets of samples before conducting data investigation is extremely difficult.   
This situation is easily noticeable when the given data analysis architectures are supposed to produce characterization of physical, chemical and biological processes. 
In this case, it is worth noting that organizing and conducting in-situ measurements is exceptionally difficult, either for ethical motivations (e.g., collecting biomedical samples from clinical trials) or for practical reasons (e.g. labeling environmental effects on large land and sea areas) \cite{multimod3,multimod8,multimod4}. 
Thus, when scarce training sets (either in terms of quantity or quality of the samples) are available, the analysis can be affected by overfitting on one hand, or low accuracy on the other \cite{multimod2,multimod9}. 

Finally, data driven data analysis structures that aim to obtain high performance accuracy results over complex datasets (such as those delivered by multimodal acquisitions, as previously described) might employ subspace reduction techniques to control the computational complexity \cite{multimod1,multimod9}. 
In this case, the original multidimensional data are mapped onto lower dimensional spaces, where linear data processing is operated to enhance efficiency. 
This approach has been proven to be successful in several cases.
However, it also shows strong degradation of performance when high variability in the data and low separability in the semantics of the considered experiments are present \cite{multimod2,multimod9,multimod6}. 
In this case, the mapping operation might not be adequate for all the samples in the dataset, given the intrinsic complexity of the records. This therefore decreases the 
 ability of the given system to generalize its characterization of the data and consequently the understanding of the processes (physical, environmental, biological, sociological) underlying the acquired records \cite{multimod1,capacity,multimod9}. 

At this point, it is worth noting that the aforementioned issues in dealing with the automation-accuracy-complexity trade-off affect architectures based on deep learning structures as well \cite{multimod2}. 
Although deep learning schemes are able to identify hidden patterns within the data, they rely on modular operations, that in turn should be expressed in terms of linear metrics, projecting operations, and/or wide processing networks. 
In order to overcome these issues, deep learning schemes for transfer learning based on graph representation of the data have recently been proposed \cite{GEOMDL}. 
However, in order to deal with non Euclidean data spaces, they rely on quantification of the informativity and relevance of the samples based on linearized distances, which may not be the best choice when dealing with datasets with high variability, low separability, or multiple data conditions across the samples, as previously motivated.

In this work, we introduce a new architecture for data analysis that aims to provide automatic and accurate data analysis in ensemble learning and transfer learning fashion, while guaranteeing low computational complexity. 
Specifically, we aim to tackle the automation-accuracy-efficiency trade-off by leveraging on \textit{adaptivity}. 
Our approach relies on understanding the relevant features that can be used to describe each sample and to run the interpretation of the results. 
This result is obtained by processing the datasets by an unsupervised scheme for dimensionality reduction based on double graph Laplacian representation of the data. 
Specifically, a joint graph based on connections weighted according to local (Gaussian kernel) and global (mutual information) metrics is employed to adaptively cluster the most significant features for each sample. 
In this way, we can identify a specific subset of features for each sample that are consequently analyzed, so to improve the informativity and relevance of the attributes to be investigated. 
These subsets are in fact analyzed alternatively by ensemble learning (at data- and model-level) and transductive transfer learning algorithms, so to overcome the issues that the traditional approaches based on these tasks face. 
To this aim, adaptive classification and non-Euclidean metrics are employed, marking a strong difference with respect to the methods that have been proposed in technical literature so far. 
Hence, the novelty of the proposed system can be described as follows:
\begin{itemize}
	\item a novel method based on adaptive dimensionality reduction for investigation of graphs induced by the data is introduced. This approach enhances the performance of state-of-the-art of dimensionality reduction architectures (e.g., spectral clustering);
	\item the classification is performed over subsets of samples with supports in the feature space that can be non coherent and non overlapping, for which non Euclidean metrics to compare the data have been employed;
	\item 
	%the computational complexity of the overall data analysis framework is reduced by 
	adaptively reducing dimensionality \textit{and} representing the data on a non-Euclidean space increases the generalization ability of the system to complex datasets where nonlinear effects (induced by high variability and low separability) are strong. 
\end{itemize}
The main characteristics of the proposed approach are reported in the next Section.

%WHAT ABOUT DEEP LEARNING?

\section{Methods}
\label{secmeth}

\begin{figure*}[htb]
	\centering
	\includegraphics[scale = 0.5]{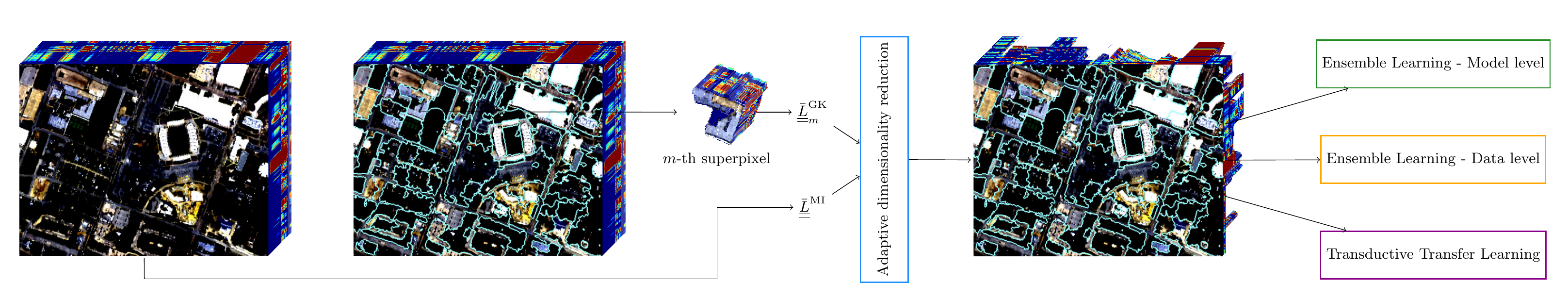} 
	\caption{Workflow of the proposed approach. The graphics are taken by considering the dataset in Section \ref{sec_exp_RS}.}
	\label{fig_workflow}
\end{figure*}

In this Section, the main steps of our proposed approach are reported. 
Specifically, we first introduce the main aspects of the adaptive dimensionality reduction strategy we propose, as well as a brief summary of the major motivations of spectral clustering, on which our approach relies on.
Then, we provide a description of the schemes of ensemble learning and transductive transfer learning that we employed to improve information extraction in multimodal data analysis. 

\subsection{Adaptive dimensionality reduction}

\subsubsection{Notation and main objective}
As previously mentioned, the method we propose to address dimensionality reduction relies on graph representation of the considered datasets.
To fully understand the main properties of our approach, let $\underline{\underline{X}} = \{x_{ln}\}_{(l,n) \in \{1, \ldots, P\} \times \{1, \ldots, N\}}$ be a dataset of $P$ samples characterized by $N$ features living in $\mathbb{R}$ across $M$ modalities. 
Thus, we consider that the records acquired by each modality are stacked columnwise in $\underline{\underline{X}}$ according to a given criterion, e.g., sequentially. 
Accordingly, the $l$-th row of $\underline{\underline{X}}$ collects the available features for sample $l$. 
Throughout the paper $\underline{x}_{*n}$ will
represent the $n$-th column of $\underline{\underline{X}}$, that corresponds to the $n$-th attribute, so it is possible to write $\underline{\underline{X}} = [\underline{x}_{*1}, \dots, \underline{x}_{*N}]$. Analogously, we denote the $l$-th row of $\underline{\underline{X}}$, that details the values of attributes at the $l$-th sample, by $\underline{x}_{l*}$, hence $\underline{\underline{X}} = [\underline{x}_{1*}^T, \dots, \underline{x}_{L*}^T]^T$. 

At this point, we can state that our adaptive dimensionality reduction scheme aims to find for a given sample $l$ the smallest subset of $K_l$ attributes $\Omega_l \subseteq \underline{x}_{l*}$, that preserves the structure and information content of the original set. 
For completeness, we can say that for two samples $i$ and $j$, with $i\neq j$, $|\Omega_i \cap \Omega_j| \in [0,\min\{K_i,K_j\}]$.

To perform such selection we apply the graph theory \cite{Luxburg} since graphs are a natural way to represent various types of data. 
The basics of graph-based clustering for feature selection are introduced in the next Section.

\subsubsection{Elements of spectral clustering}
Representing complex datasets as graph structures identifies a solution that has been widely employed in several research fields to detect and understand hidden patterns among the considered records. 
Indeed, several applications - from Earth science to biomedical signal processing, from computer vision to molecular chemistry - have taken advantage  of the ability of graph theory to describe sophisticated interactions and relationships among samples in a consistent way, which allows robust and accurate investigation of the datasets \cite{Luxburg,specclust1,specclust2,specclust3}.
Moreover, by exploring the considered data in a topological domain, graph theory-based methods show high versatility, so that multiple sources of information can be properly characterized. 
These properties are achieved by means of well defined algebraic operations, which allow the implementation of sophisticated data processing steps in linear programming \cite{Luxburg}. 

When applied to dimensionality reduction, spectral clustering plays a key role within graph theory-based approaches \cite{specclust1,specclust2,specclust3}. 
Clustering enables an accurate discrimination of the most relevant characteristics among the samples the considered dataset consists of.
Specifically, graph-based clustering aims to divide the graph induced by the considered dataset into subgraphs: each subgraph is meant to group together the features that are considered to be similar according to a given criterion and/or metric. 
In this way, graph-based clustering enables the reduction of redundancy within the considered dataset (i.e., only the features that are considered informative are kept for subsequent analysis steps) \cite{specclust2}. 
It also aims to address data variability, since the fluctuations of the attributes caused by non-idealities and inconsistency should be leveraged by the analysis of the statistical and spectral properties of the features \cite{specclust1}. 

Let us take a closer look to the main steps of spectral clustering for dimensionality reduction.
First of all, this approach requires to determine a graph induced by the features of the considered dataset \cite{specclust1,specclust2,specclust3}. 
In detail, spectral clustering methods work on a graph $\mathbb{G}({\cal V}, {\cal E})$ that consists of a set of vertices (or nodes) ${\cal V}$ which are connected by a set of edges ${\cal E}$. 
Thus, according to the notation that has been previously introduced in this Section, spectral clustering methods assume that each vertex $v_i$ in ${\cal V}$ identifies the $i$-th feature of the dataset $\underline{\underline{X}}$. 
Moreover, two vertices $v_i$ and $v_j$ are connected by an edge ${\cal E}_{ij}$ that is characterized by a weight $w_{ij}$. 
The value of this quantity is determined by the similarity metric that is considered in the clustering set-up. 
The metric that is typically used to define the weight of ${\cal E}_{ij}$ is based on the Gaussian kernel distance (or radial basis function), so that we can write $w_{ij}$ as follows:

\begin{eqnarray} \label{eq_specclust1}
w_{ij} &=& \exp \left[ \frac{-||\underline{x}_{*i} - \underline{x}_{*j}||^2}{2\sigma}\right] ,
\label{eq_GKMI0}
\end{eqnarray}

\noindent where $\sigma$ controls the width of Gaussian kernel. 
%the neighborhood in the graph, i.e., the number of connected vertices. 
 
The graph $\mathbb{G}$ can be then described by a matrix $\underline{\underline{W}}$ whose elements are all the weights $w$, i.e., $\underline{\underline{W}} = \{w_{ij}\}_{(i,j)\in \{1, \ldots, N\}^2}$ \cite{Luxburg}. 
$\underline{\underline{W}}$ is called the adjacency matrix of $\mathbb{G}$. 
At this point, 
we can define $d_i = \sum_{j=1}^{N} w_{ij}$ as the degree of the $i$-th vertex of $\mathbb{G}$. 
Consequently, it is possible to summarize these quantities in a vector $\underline{d} = [d_i]_{i=1, \ldots, N}$. 
The degree matrix of $\mathbb{G}$ can be then defined as $\underline{\underline{D}} = \text{diag}(\underline{d})$, i.e., the $(i,j)$ element of $\underline{\underline{D}}$ is equal to $d_i$ if $i=j$, whereas it is set to 0 if $i \neq j$. 
We can define the Laplacian matrix of $\mathbb{G}$ as $\underline{\underline{L}} = \underline{\underline{D}} - \underline{\underline{W}}$. 
Finally, the normalized Laplacian matrix of $\mathbb{G}$ is defined as $\bar{\underline{\underline{L}}} = \underline{\underline{D}}^{-1/2} \underline{\underline{L}} \underline{\underline{D}}^{-1/2} =  \underline{\underline{I}} - \underline{\underline{D}}^{-1/2} \underline{\underline{W}} \underline{\underline{D}}^{-1/2}$, being $\underline{\underline{I}}$ the $N \times N$ identity matrix. 

The characteristics of $\underline{\underline{L}}$ and $\bar{\underline{\underline{L}}}$ are very relevant to understand the properties of the considered graph \cite{Luxburg}. 
In particular, the eigenanalysis of these matrices can be used to efficiently identify the most similar groups of vertices that show up in the graph, i.e., the number and type of connected components in $\mathbb{G}$ \cite{specclust1,specclust2}. 
Specifically,  
the multiplicity K of the eigenvalue 0 of the Laplacian matrix equals the
number of connected components $C_\kappa$, $\kappa = 1, \ldots, K$ in the graph. 
Thus, the vertex set ${\cal V}$ would result to be partitioned in the $C_\kappa$ subsets. 
The eigenspace of eigenvalue 0 is then spanned
by the indicator vectors $\underline{h}_\kappa$,  $\kappa = 1, \ldots, K$ of those components. 
The indicator vectors can be organized in a matrix $\underline{\underline{H}} = [\underline{h}_\kappa^T]_{\kappa = 1, \ldots, K}$, where $\underline{h}_\kappa = [h_{\kappa n }]_{n \in \{1,\ldots,N\}  }$ and $\underline{h}_\kappa^T$ is its transpose. 
Thus,  $\underline{\underline{H}} \in \mathbb{R}^{N \times K}$. 
Indeed, the value of each $h_{\kappa n}$ is set to the inverse of the cardinality of the connected component $C_\kappa$ if vertex $n$ belongs to it, otherwise it is set to 0. 
In other terms, $h_{\kappa n} = 1/\sqrt{|C_{\kappa}|}$ if $v_n \in C_\kappa$, whereas $h_{\kappa n} = 0$ otherwise, so that $\underline{\underline{H}}^T \underline{\underline{H}} = \underline{\underline{I}}$  \cite{Luxburg}. 

Therefore, the eigenanalysis of the Laplacian matrix of an undirected graph can provide information on the relevant features the given dataset consists of \cite{specclust1,specclust3}.
Specifically, this investigation aims to identify the groups of features that behave coherently across the data, hence implicitly addressing the redundancy of non-relevant features and pointing out the presence of corrupted and/or highly perturbated variables across the records. 
This property is used by most of the spectral clustering techniques applied to feature selection that have been proposed in technical literature. 
Once the eigenvectors of $\underline{\underline{L}}$ (or $\bar{\underline{\underline{L}}}$) have been computed, schemes using clustering techniques based on k-means approach or more sophisticated algorithms have been proposed \cite{specclust1,specclust2,specclust3}. 

In this respect, one of the most utilized approaches relies on the equivalence of the identification of the indicator vectors that have been previously mentioned with the normalized graph cut problem \cite{specclust1,specclust3}. 
In fact, 
the main idea behind spectral clustering is to separate points in different groups according to their similarities. 
On the other hand, for
data given in the form of a similarity graph, 
this problem is functionally equivalent to finding a partition
of the graph such that the edges between different groups have a very low weight  and the edges within a group have a high
weight: in the first case, points in different clusters are dissimilar from each other; in the second case, points within the same cluster are similar to each other \cite{Luxburg,specclust2}. 
This problem has been stated and solved in terms of graph segmentation by means of a global criterion for graph partitioning \cite{specclust3}. 
Specifically, the problem of partitioning a graph in K connected components $C_\kappa$, $\kappa=1, \ldots, K$, can be solved by minimizing over $\{C_\kappa\}_{\kappa = 1, \ldots, K}$ the RatioCut function $RC_K$, which can be written as follows:

\begin{eqnarray}
RC_K &=& \frac{1}{2} \sum_{\kappa=1}^{K} \frac{\zeta(C_\kappa,\bar{C}_\kappa)}{\tilde{C}_\kappa}, 
\label{eq_specclust2}
\end{eqnarray} 

\noindent where $\bar{C}_\kappa$ represents the complement of the $\kappa$-th partition over the vertex set ${\cal V}$, $\tilde{C}_\kappa$ is a measure of the width of the $\kappa$-th partition (either in terms of cardinality or volume), and $\zeta(C_\kappa,\bar{C}_\kappa) = \sum_{v_i \in C_\kappa, v_j \in \bar{C}_\kappa} w_{ij}$. 
By minimizing $RC_K$, the weights of the edges connecting the nodes within $C_\kappa$ will be large, while the edges connecting nodes within $C_\kappa$ with the nodes in its complement $\bar{C}_\kappa$ will be small. 
At this point, it is worth recalling that it has been proven that the following equation holds \cite{Luxburg,specclust2}:

\begin{eqnarray}
RC_K =& \sum_{\kappa=1}^{K} \underline{h}_\kappa^T \bar{\underline{\underline{L}}} \underline{h}_\kappa = & \text{Tr}( \underline{\underline{H}}^T \bar{\underline{\underline{L}}} \underline{\underline{H}}), 
\label{eq_specclust3}
\end{eqnarray}

\noindent where $\text{Tr}(\cdot)$ is the trace operator. 
Thus, the problem of minimizing the graph cut can be written as follows: 

\begin{eqnarray}
\min_{\{C_\kappa\}_{\kappa=1,\ldots,K}} \text{Tr}( \underline{\underline{H}}^T \bar{\underline{\underline{L}}} \underline{\underline{H}}) & \text{subject to} &  \underline{\underline{H}}^T \underline{\underline{H}} = \underline{\underline{I}}.
\label{eq_specclust4}
\end{eqnarray}

It has been proven that 
the solution to this trace minimization problem is given by choosing
$\underline{\underline{H}}$ as the matrix which contains the first K eigenvectors of $\bar{\underline{\underline{L}}}$ as columns. 
This point means that the problem in (\ref{eq_specclust4}) can be efficiently solved by linear programming, and consequently makes spectral clustering one of the most appealing techniques for dimensionality reduction in data analysis community \cite{specclust1,specclust2,specclust3}. 

\subsubsection{Issues in classic spectral clustering and proposed approach} 
\label{sec_meth_ADR}
Spectral clustering offers an attractive
alternative which clusters data using data-driven eigenanalysis, since it is based on a similarity matrix that can be derived from the data directly \cite{specclust3,specclust2}. 
As long as pair-wise similarity metrics can be defined on the records, spectral clustering can be applied, while other dimensionality reduction and/or clustering methods might not. 
In fact, in an ideal case - in which all points in different clusters are infinitely far
apart - it would be possible to separate the data into different groups by taking advantage of the high degree of multiplicity of the null eigenvalue in the Laplacian matrix induced by the data. 
Therefore, by taking advantage of a data-driven approach on the algebraic properties of the similarities among data, the informative eigenvectors would appear with distinctive gaps, which can be used to perform an accurate feature selection \cite{Luxburg}. 

However, there are still several issues in spectral clustering that might limit its use in several application scenarios and especially in operational contexts \cite{specclust1,specclust2}. 
Specifically, 
 when analyzing large scale heterogeneous datasets (such as those to be taken into account when addressing operational problems), the records are typically characterized by non uniform geometrical, spectral, and statistical distributions across the data to be explored. 
The complexity of the data reverberates in the graph representation \cite{Luxburg,mackaybook,GEOMDL}. 
In terms of graph theory, this leads to graph representations typically characterized by fully connected topologies, which are hard to investigate, especially for discrimination purposes. 
Furthermore, using aggregate metrics  (such as the one in (\ref{eq_specclust1})) might not be the best choice when records semantically belonging to the same thematic cluster show up with high feature variability caused by varying conditions in the acquisitions across the whole considered dataset \cite{gulikers17, COUILLET18, COUILLET20}. 
This is particularly evident in multimodal data analysis, where each modality might assume different levels of significance and/or informativity across neighborhoods of samples and features \cite{multimod1,multimod2,Luxburg}. 

It is worth to recall the sum of the
elements of each row of the Laplacian matrix
is zero by construction \cite{CommunityDetection_Fortunato,Luxburg}. 
From a practical point of view, this means that the multiplicity of the null eigenvalue is 1 \cite{COUILLET20,COUILLET18}, corresponding to a constant eigenvector \cite{specclust1,Luxburg}. 
In fact, the null eigenvalues' multiplicity refers to the number of connected components of a graph \cite{Luxburg}. While this number represents the clusters of a sparse graph, where the subgraphs are disconnected, it will only refer to the graph itself when it is fully connected\cite{CommunityDetection_Fortunato}. In the latter case, the next lowest eigenvalues are indicators of the clusters \cite{CommunityDetection_Fortunato,Luxburg}. 
However, in the case of large scale heterogeneous data that can be characterized by sparsity (such as multimodal datasets),  the spectrum of the remaining eigenvalues would be pretty flat \cite{COUILLET18,COUILLET20}, so that understanding their relevance (to discriminate the graph indicators that are actually carrying the information on the significant features in the data) would be jeopardized. 
In fact, in a realistic operational case where noise and a fair amount of similarities between clusters exists, the
distribution of elements of an eigenvector is extremely complex, and the gaps between clusters
in the elements of the top eigenvectors are blurred \cite{specclust3}.
Hence, it is hard to understand how many connected components the graph could be partitioned in, and it is hard to point out the 
eigenvectors that are uninformative. 
This would be a major drawback in the data analysis chain, as this reflects to selecting non-relevant features for subsequent processing steps, allowing possible degradation of the analysis \cite{Luxburg,multimod1,ELdata_pruning5}.

It is therefore possible to identify the following issues that are still largely unsolved in technical literature:

\begin{itemize}
	\item how to automatically determine the number of clusters; 
	\item how to perform effective clustering when sparse
	data are considered;
	\item how to determine a similarity metric that could take into account the variability of the data (especially when records acquired by heterogeneous sensors and platforms are taken into account). 
\end{itemize}

It is true that several methods have been proposed in technical literature to address these points. 
However, these schemes struggle to solve these issues simultaneously \cite{Luxburg,specclust1,specclust2,specclust3}. 

In this work, we propose tackling the aforesaid issues by  exploiting the ability of spectral clustering to explore graph connectivity.
First, we address adaptivity in our scheme by grouping the original dataset $\underline{\underline{X}}$ into local neighborhoods of samples showing homogeneous characteristics across the records. 
This step allows us to consider the multiple characteristics of the samples that each multimodal dataset consists of, and appropriately take into account their variability in distilling their true informativity across the data. 
In order to group samples showing homogeneous properties from a statistical and geometrical point of view, we partition the datasets in clusters of samples $\{{\cal S}_m\}_{m=1, \ldots,S}$ by means of the watershed  approach \cite{watershed}. 
This scheme aims to identify supersamples with similar distributions across the considered dataset by computing the gradient vector field in the feature space from samples showing the highest energy. 
In this way, it is possible to obtain a reliable partition of the original dataset in local neighborhoods with similar properties.   
As a result, we can write $\underline{\underline{X}} = \bigcup_{m=1}^M {\cal S}_m$, ${\cal S}_l \cap {\cal S}_m = \emptyset$ $\forall(l,m) \in \{1, \ldots, M \}^2$, $l \neq m$. 
Further, the cardinality of each ${\cal S}_m$ might differ for each $m$, i.e., ${\cal S}_m = \{ x_{m_{tn}} \}_{(t,n) \in \{1, \ldots, P_m\} \times \{1, \ldots, N\}}$. 
Thus, $\underline{x}_{m_{*n}}$ identifies the vector of the $n$-th feature acquired across the samples that are part of ${\cal S}_m$. 
%Since watershed approach has been mainly applied to image processing tasks, we can refer to the $m$-th sample group ${\cal S}_m$ as the $m$-th superpixel.  
Throughout the paper we will refer to ${\cal S}_m$ as the $m$-th supersample. 

In order to improve the identification of the most relevant features across $\underline{\underline{X}}$, we will construct a graph for every supersample in the feature graph connecting all vertices according to the similarities we can compute in each ${\cal S}$. 
In order to improve our ability to select informative features in the supersample (although sparsity and high variability that can occur in operational scenarios), we build a graph where the two edges connect each vertex. 

In other terms, for the supersample ${\cal S}_m$ we build the graph $\mathbb{G}_m=({\cal V},{\cal E}_m^{\text{GK}}, {\cal E}^{\text{MI}})$. 
The $n$-th vertex $v_n$ in ${\cal V}$ identifies the $n$-th feature each sample in ${\cal S}_m$ is described by. 
Two vertices $v_{n_1}$ and $v_{n_2}$ are connected by two kinds of edges, ${\cal E}_{m_{n_1 n_2}}^{\text{GK}}$ and ${\cal E}_{{n_1 n_2}}^{\text{MI}}$. 
The edge ${\cal E}_{m_{n_1 n_2}}^{\text{GK}}$ is characterized by a weight that is computed across the $m$-th supersample according to the Gaussian kernel function. 
Specifically, we can define the weight of ${\cal E}_{m_{n_1 n_2}}^{\text{GK}}$ as follows: 

\begin{eqnarray}
w_{m_{n_1 n_2}}^{\text{GK}} & =&  \exp \left[ -\frac{|| \underline{x}_{m_{*n_1}}- \underline{x}_{m_{*n_2}} ||^2}{2\sigma} \right] . \label{eq_GKMI1}
\end{eqnarray}

The value of $\sigma$ controls the width of the Gaussian kernel as in (\ref{eq_GKMI0}). 
With this in mind, if the features $n_1$ and $n_2$ are very similar across the $m$-th supersample, the value of $w_{m_{n_1 n_2}}^{\text{GK}}$ will be large, and it is possible to assume that considering just one of them could be enough to achieve an accurate characterization of the supersample. 
On the other hand, a small value of  $w_{m_{n_1 n_2}}^{\text{GK}}$
would indicate that the two features are carrying different information, thus they should be both taken into account to run the analysis of the dataset \cite{Luxburg}. 
Therefore, this similarity measure based on the Gaussian kernel represents
the structure of the feature set. In our method, it
is applied at local level (i.e., on each supersample) in order
to preserve the local particularities of the original data.  

This approach might help us in overcoming the variability of the statistical properties of the data across the complete dataset, being able to grasp the local characteristics of each feature. 
Nevertheless, this might not be enough to address the redundancy and the interplay among features we could record across the dataset, especially in case of complex datasets (leading to fully connected graph representations) as previously mentioned \cite{ELdata_pruning5}. 
In order to address this point, we consider connecting each vertex by another kind of edge, whose weights have to be computed across the whole dataset $\underline{\underline{X}}$. 
Specifically, we compute their values in terms of mutual information between the features, so to evaluate the degree of redundancy and intercorrelation between them \cite{mackaybook,MUTINF6}. 
Thus, we can write the weight of ${\cal E}_{n_1 n_2}^{\text{MI}}$ in terms of mutual information between the two features as follows:
 
\begin{eqnarray}
w_{n_1 n_2}^{\text{MI}} =& \sum_{i = 1}^P\sum_{j = 1}^P P(x_{in_1},x_{jn_2})\log \frac{P(x_{in_1},x_{jn_2})}{P(x_{in_1})P(x_{jn_2})}, \label{eq_GKMI2}	
\end{eqnarray} 
%\begin{eqnarray}
%w_{n_1 n_2}^{\text{MI}} &=& I(\underline{x}_{*n_1}, \underline{x}_{*n_2}), \nonumber
%\\
% &= & \sum_{i = 1}^P\sum_{j = 1}^P P(x_{in_1},x_{jn_2})\log\left( \frac{P(x_{in_1},x_{jn_2})}{P(x_{in_1})P(x_{jn_2})}\right) \, \label{eq_GKMI2}	  
%\end{eqnarray}

\noindent where $1\leq n_1,n_2\leq N$, $P(y,z)$ is the joint density function of $y$ and $z$, and $P(z)$ is marginal of $z$. Mutual information quantifies the shared information between two random variables \cite{MUTINF6}. Accordingly, large values of $w_{n_1 n_2}^{\text{MI}}$
%$I(\underline{x}_{*n_1}, \underline{x}_{*n_2})$ 
imply redundancy in information. On the other hand, low values of $w_{n_1 n_2}^{\text{MI}}$
%$I(\underline{x}_{*n_1}, \underline{x}_{*n_2})$ 
imply synergy (novelty). 

At this point, we can define the adjacency matrices according to the quantities in (\ref{eq_GKMI1}) and (\ref{eq_GKMI2}) as $\underline{\underline{W}}_m^{\text{GK}} = \{w_{m_{ij}}^{\text{GK}}\}_{(i,j)\in \{1, \ldots, N\}^2}$ and $\underline{\underline{W}}^{\text{MI}} = \{w_{ij}^{\text{MI}}\}_{(i,j)\in \{1, \ldots, N\}^2}$, respectively. 
Accordingly, we can define the degree matrices $\underline{\underline{D}}_m^{\text{GK}} = 
\text{diag}(\underline{d}_m^{\text{GK}})$ and 
$\underline{\underline{D}}^{\text{MI}} = \text{diag}(\underline{d}^{\text{MI}})$, where $\underline{d}_m^{\text{GK}} = [d_{m_n}^{\text{GK}} = \sum_{j=1}^{N} w_{m_{nj}}^{\text{GK}}]_{n=1, \ldots, N}$
and $\underline{d}^{\text{MI}} = [d_n^{\text{MI}} = \sum_{j=1}^{N} w_{nj}^{\text{MI}}]_{n=1, \ldots, N}$. 
Consequently, we can define the normalized Laplacian matrix associated with $\underline{\underline{W}}_m^{\text{GK}}$ and
$\underline{\underline{W}}^{\text{MI}}$ as $\bar{\underline{\underline{L}}}_m^{\text{GK}} = \underline{\underline{I}} -  \underline{\underline{D}}_m^{\text{GK}^{-1/2}}
\underline{\underline{W}}_m^{\text{GK}}
\underline{\underline{D}}_m^{\text{GK}^{-1/2}}$
 and
 $\bar{\underline{\underline{L}}}^{\text{MI}} = \underline{\underline{I}} -  \underline{\underline{D}}^{\text{MI}^{-1/2}}
 \underline{\underline{W}}^{\text{MI}}
 \underline{\underline{D}}^{\text{MI}^{-1/2}}$. 
 
Our goal is to partition the graph such that the vertices of the same subgraph have strong connections via both links, while the vertices from different subgraphs have one or two weak connections. 
According to the notation and the main findings reported in the previous Section, 
this problem can be written as follows

\begin{eqnarray}
\begin{cases}
\min_{\underline{\underline{H}}} & \text{Tr}( \underline{\underline{H}}^T \bar{\underline{\underline{L}}}_m^{\text{GK}}  \underline{\underline{H}}) %\nonumber
\\ %& \\
%& \text{subject to} \quad \underline{\underline{H}}^T \underline{\underline{H}} = \underline{\underline{I}} \\
 \min_{\underline{\underline{H}}}  & \text{Tr}( \underline{\underline{H}}^T \bar{\underline{\underline{L}}}^{\text{MI}}  \underline{\underline{H}}) %\label{eq_GKMI3} %&
%\min_{\H} \Tr\left(\H\L_l^\gk\H\right) &\\
%&\text{subject to} \quad \H^T\H = \I\\
%\min_{\H} \Tr\left(\H\L_l^\mi\H\right) &
\end{cases}
\label{eq_GKMI3}
\end{eqnarray}

%\begin{equation*}
%\left \{ \begin{array}{rl}
%	\min_{\underline{\underline{H}}}  \text{Tr}( \underline{\underline{H}}^T \bar{\underline{\underline{L}}}_m^{\text{GK}}  \underline{\underline{H}})\\
%	\min_{\underline{\underline{H}}}  \text{Tr}( \underline{\underline{H}}^T \bar{\underline{\underline{L}}}_m^{\text{MI}}  \underline{\underline{H}})
%\end{array}
%\right.
%\label{eq_GKMI3}
%\end{equation*}

\noindent subject to $\underline{\underline{H}}^T \underline{\underline{H}} = \underline{\underline{I}}$. 
The solution of (\ref{eq_GKMI3}) is given by the common
eigenspace of $\bar{\underline{\underline{L}}}_m^{\text{GK}}$
 and $\bar{\underline{\underline{L}}}^{\text{MI}}$  , i.e., their joint eigenvectors.  %which can be identified by solving the following:  
 Hence, problem (\ref{eq_GKMI3}) turns into a \emph{joint diagonalization problem} that seeks the matrix ${\underline{\underline{V}}_m}$ such that the products $\underline{\underline{V}}_{m}^T\bar{\underline{\underline{L}}}_m^{\text{GK}} \underline{\underline{V}}_{m}$ and $\underline{\underline{V}}_{m}^T\bar{\underline{\underline{L}}}^{\text{MI}} \underline{\underline{V}}_{m}$ are as diagonal as possible. Among the diagonality criteria proposed in the literature, we consider the criterion proposed by Pham, which is given as the Kullback-Leibler divergence between the previously mentioned products and their diagonal forms\cite{Pham}. As such the common eigenvectors $\bar{\underline{\underline{L}}}_m^{\text{GK}}$
 and $\bar{\underline{\underline{L}}}^{\text{MI}}$ can be identified by solving the following \cite{Ablin2018}:
%The
%common eigenspace spanned by both Laplacians enable their
%interaction which might unfold complicated structure of the
%graph. 
%The joint eigenvectors of the graph Laplacians $\bar{\underline{\underline{L}}}_m^{\text{GK}}$
%and $\bar{\underline{\underline{L}}}^{\text{MI}}$ are defined so that the following equations hold:
%\begin{align}
%\bar{\underline{\underline{L}}}_m^{\text{GK}} &= \underline{\underline{V}}_m \underline{\underline{\Lambda}}_m^{\text{GK}} \underline{\underline{V}}_m^{T}, \label{eqn:eigen_gk}\\ 
%\bar{\underline{\underline{L}}}^{\text{MI}}  &= \underline{\underline{V}}_m \underline{\underline{\Lambda}}^{\text{MI}}
%\underline{\underline{V}}_m^{T},
%\label{eqn:eigen_mi}
%\end{align} 
%
%\noindent where $\V_{l} = [\v_{l1}, \dots, \v_{lN}]$ is the matrix of eigenvectors, and $\mathbf{\Lambda}_{l}^{\gk} = \diag(\lambda_{l1}^{\gk},...,\lambda_{lN}^{\gk})$ and $\mathbf{\Lambda}_{l}^{\mi} = \diag(\lambda_{l1}^{\mi},...,\lambda_{lN}^{\mi})$ are diagonal matrices of the corresponding GK- and MI-based eigenvalues, respectively. 
%This goal can be achieved by means of a joint diagonalization of the eigenspace spanned by both Laplacians, which can be described as follows: 

\begin{eqnarray} 
\min_{\underline{\underline{V}}_m}
\log  \frac{\left|\text{diag}(\underline{\underline{V}}_{m}^T\bar{\underline{\underline{L}}}_m^{\text{GK}} \underline{\underline{V}}_{m})\right|}{\left|\underline{\underline{V}}_{m}^T\bar{\underline{\underline{L}}}_m^{\text{GK}} \underline{\underline{V}}_{m}\right|}  
+ \log \frac{\left|\text{diag}(\underline{\underline{V}}_{m}^T\bar{\underline{\underline{L}}}^{\text{MI}} \underline{\underline{V}}_{m})\right|}{\left|\underline{\underline{V}}_{m}^T\bar{\underline{\underline{L}}}^{\text{MI}} \underline{\underline{V}}_{m}\right|},
\label{eqn:criterion}
\end{eqnarray} 

\noindent where $\underline{\underline{V}}_m = [\underline{v}_{m1}, \dots, \underline{v}_{mN}]$ is the matrix of the joint eigenvectors. 
The matrix $\underline{\underline{H}}$ in (\ref{eq_GKMI3}) contains the eigenvectors corresponding to the  $K_m$ lowest and non-null eigenvalues.
At this point, 
the original set of features can be embedded in a lower dimensional manifold by a $\underline{\underline{H}}$-based projection \cite{Luxburg,specclust1}. 
In this space, it is possible to use a classic $K_m$-means to partition the embedding, where $K_m$ (i.e., the number of relevant features in the $m$-th supersample) is chosen by means of the kneedle algorithm \cite{kneedle}. 
The number of relevant features equals the number of informative eigenvalues that the kneedle algorithm determines as the local minima of the eigenvalues' difference curve , if the eigengap with the remainder spectra is pronounced \cite{kneedle}. 
Finally, the  centroids of the clusters will form the set of selected attributes $\Omega_m$ for the samples in supersample $m$. 
This new representation enhances the classic spectral clustering in several aspects, such as:
\begin{itemize}
	\item the common eigenspace spanned by both Laplacians enable their interaction which might unfold the complicated structure of the
	graph; 
	\item the effectiveness of classic spectral clustering approaches is improved by increasing the separability of data, mainly if it is separable in a non-linear manner; 
	\item dramatically reducing the sensitivity to the initial state of classic spectral clustering.
\end{itemize}  

Moreover, it is worth emphasizing how the proposed approach is able to smoothly solve the automatic choice of the number of features to be selected \cite{specclust1,specclust2,specclust3}. 
The proposed scheme in fact able to overcome the issue caused by the complexity of datasets acquired in operational scenarios - which are typically characterized by sparsity and high variability - which translates in a flat spectrum of the eigenvalues when classic spectral clustering strategies based on a single similarity metric are employed. 
The proposed approach enhances the separability of the eigenvectors of the Laplacian matrices, and hence 
allows us to employ a robust algorithm (such as Kneedle method \cite{kneedle}) to identify the eigenvectors associated with the smallest eigenvalues, which are representing the most relevant features in the supersample. 

\subsection{Proposed ensemble learning and transductive transfer learning}

Two main outputs resulting from the application of the aforementioned adaptive dimensionality reduction algorithm to the multimodal dataset $\underline{\underline{X}}$ can be listed as follows:

\begin{itemize}
	\item the original dataset is partitioned in $M$ groups of samples (or supersamples), i.e., $\underline{\underline{X}} = \bigcup_{m=1}^M {\cal S}_m$, ${\cal S}_l \cap {\cal S}_m = \emptyset$ $\forall(l,m) \in \{1, \ldots, M \}^2$, $l \neq m$;
	\item for the $m$-th supersample ${\cal S}_m$, a subset of $K_m$ features $\Omega_m$ is selected. Hence, for two different supersamples ${\cal S}_s$ and ${\cal S}_t$  $|\Omega_s \cap \Omega_t| \in [0,\min\{K_s,K_t\}]$. 
\end{itemize}

At this point, we proposed three different approaches to improve the information extraction in multimodal data analysis. 
These strategies are summarized as follows:

\subsubsection{Ensemble learning at model level}
\label{sec_meth_RF}
First, we aim to take advantage of the ability of the adaptive dimensionality reduction method we previously introduced by considering the various subsets of features it can select for each supersample. 
Thus, we set up an ensemble of weak classifiers  which perform classification by taking into account the features that have been selected and collected in $\{\Omega_m\}_{m=1, \ldots, M}$. 
Specifically, for each combination of features $\Omega_m$, we implement $\Lambda$ weak classifiers \cite{randforPR1,randforPR2}. 
Therefore, the overall ensemble learning architecture results in $\Lambda M $ classifiers. 
The final output is then established according to a majority rule applied to the decisions taken by each weak classifier. 
In this work, we focus our attention on the use of decision trees as weak classifiers.

\subsubsection{Ensemble learning at data level}
\label{sec_meth_SVM}
In this case, for each supersample ${\cal S}_m$, information extraction is performed by taking into account only the subset of selected features corresponding to the $m$-th supersample, i.e., $\Omega_m$. 
As in this paper we focus our attention on classification results, in this case we employ a support vector machine (SVM) algorithm to analyze the samples within ${\cal S}_m$: hence, $M$ SVMs are set up to classify the data. 
The training set is chosen across the whole dataset (i.e., considering samples that might be outside of the supersample ${\cal S}_m$), whilst the hyperplane detection is performed in the multidimensional space spanned by $\Omega_m$.

\subsubsection{Transductive transfer learning}
\label{sec_meth_TTL}
In order to perform transfer learning, we divide the original dataset $\underline{\underline{X}}$ in two separate domains, $\underline{\underline{X}}^{\cal S}$ and $\underline{\underline{X}}^{\cal T}$ \cite{TTL1}. 
The first domain, or source domain, contains $M^{\cal S}$ supersamples and their corresponding labels. 
On the other hand, the target domain $\underline{\underline{X}}^{\cal T}$ consists of supersamples where no prior knowledge of the labels is available. 
The objective is therefore to retrieve the labels for each sample in $\underline{\underline{X}}^{\cal T}$ by using the model that can be learned by $\underline{\underline{X}}^{\cal S}$. 
Specifically, the information is propagated from source to target domain in an iterative fashion by maximizing the Jaccard index between the subsets of selected features the samples in $\underline{\underline{X}}^{\cal S}$  and $\underline{\underline{X}}^{\cal T}$ share \cite{propagationJaccard}. 
It is worth recalling that the Jaccard index is a metric used to quantify the similarity of two multivariate items showing partial overlap of the feature sets they consist of. 
In this respect, Jaccard index can be considered as a non-Euclidean metric 
which can be described in terms of the volume spanned by the intersection of the simplices induced by the considered items in the multidimensional feature space. 
In this way, it is possible to obtain a reliable understanding of the actual difference between two samples with different feature supports. 
Hence, in our approach, the subset of attributes corresponding to the intersection of the attribute sets considered in the source and target subpixels is used to infer labels.
Propagation is iteratively performed by inferring the labels in the samples in $\underline{\underline{X}}^{\cal T}$ that are more similar to those in $\underline{\underline{X}}^{\cal S}$ in terms of Jaccard index \cite{TTL1,propagationJaccard}. 

\section{Experimental results $\&$ discussion}
\label{secresult}

The proposed approach has been tested over several multimodal datasets representative of different research fields. 
In this Section, we first summarize the main characteristics of the datasets we analyzed. 
%we provide a brief validation of the adaptive dimensionality reduction step, by showing how it is able to identify the most relevant features in very noisy conditions. 
%Finally, 
We then report the performance of the proposed ensemble learning and transfer learning frameworks on these datasets.

\subsection{Datasets}
\label{sec_exp_data}

We tested the proposed approach on three very diverse datasets, focusing on three different research fields: remote sensing, brain-computer interface, and photovoltaic energy. 

\subsubsection{Multimodal remote sensing}
\label{sec_exp_RS}
First, we considered the multimodal dataset that has been used for the IEEE GRSS Data Fusion Contest 2013. 
The dataset consisted of LiDAR and Hyperspectral data acquired over the University of Houston campus and the neighboring urban area, and was distributed for the 2013 IEEE GRSS Data Fusion Contest~\cite{data_houston}. Hyperspectral data was acquired from the Compact Airborne Spectrographic Imager (CASI) with 2.5 m spatial resolution. The hyperspectral dataset includes 144 spectral bands ranging from 0.38 to 1.05 $\mu{m}$. The available ground truth labels consisted of 15 classes. 
Houston dataset contains hyperspectral data (144 bands) and 1$\times$LiDAR data (including one band  and 6 textural features). The final dataset that was used consisted of $N = 151$ features over an area of 1905$\times$349 pixels.

\subsubsection{Multimodal brain-computer interface}
\label{sec_exp_BCI}
The second dataset we considered was collected by recording electroencephalography
(EEG), electromyography
(EMG) and electro-oculography (EOG) data on 11 intuitive upper extremity
movements from 25 participants by means of bran-computer interface \cite{BrainComputer1}. 
The data, acquired using a 60-channel EEG, 7-channel EMG, and
4-channel EOG, were simultaneously recorded during  a 3-session experiment, collecting 3300 trials per participant. 
According to the notation we have used in Section \ref{secmeth}, the final dataset consists of $N=71$ multimodal features for a total sum of 82500 samples across all participants. 

\subsubsection{Multimodal photovoltaic energy}
\label{sec_exp_PV}
Finally, we analyzed the data collected at the University of Queensland, Australia, for spatiotemporal monitoring of the photovoltaic energy produced between July 21 and August 17, 2018 \cite{data_PhotoVoltaic}. 
For each day, 1440 samples acquired by the weather ground stations (average and instantaneous wind speed [km/h] and direction [deg], temperature [deg], relative humidity [\%], mean surface level pressure [hPa], accumulated rain [mm], rain intensity [mm/h], accumulated hail [hits/cm$^2$], hail intensity [hits/cm$^2$h], solar mean [W/m$^2$]) have been collected. 
For each sample, the photovoltaic energy [W/h] is recorded: 10 classes uniformly drawn based o this parameter are considered. 
The considered data analysis task comprised assigning a class to all the 1440 $\times$ 28  samples  by investigating the $N=11$ heterogeneous features. 

%\color{red}
%\subsection{Validation of adaptive dimensionality reduction strategy}

\color{black}

\subsection{Results}

\subsubsection{Ensemble learning at model level} 
\label{sec_exp_RF}
Let us consider the proposed architecture for ensemble learning that has been described in Section \ref{sec_meth_RF}. 
As previously mentioned, we analyzed the considered dataset by means of an ensemble of weak classifiers (decision trees) initialized according to the feature subsets $\Omega$ that have been identified by means of the adaptive dimensionality reduction (ADR) strategy that has been introduced in Section \ref{sec_meth_ADR}. 
We compared the achieved results in terms of classification accuracy with the outcomes delivered by two of the most relevant ensemble learning state-of-the-art techniques that have been introduced in technical literature, gradient boosting and random forest. 
We carried out 100 experiments for varying quantities of weak classifiers. 

\begin{figure}[htb]
	\centering
	\includegraphics[width=1\columnwidth]{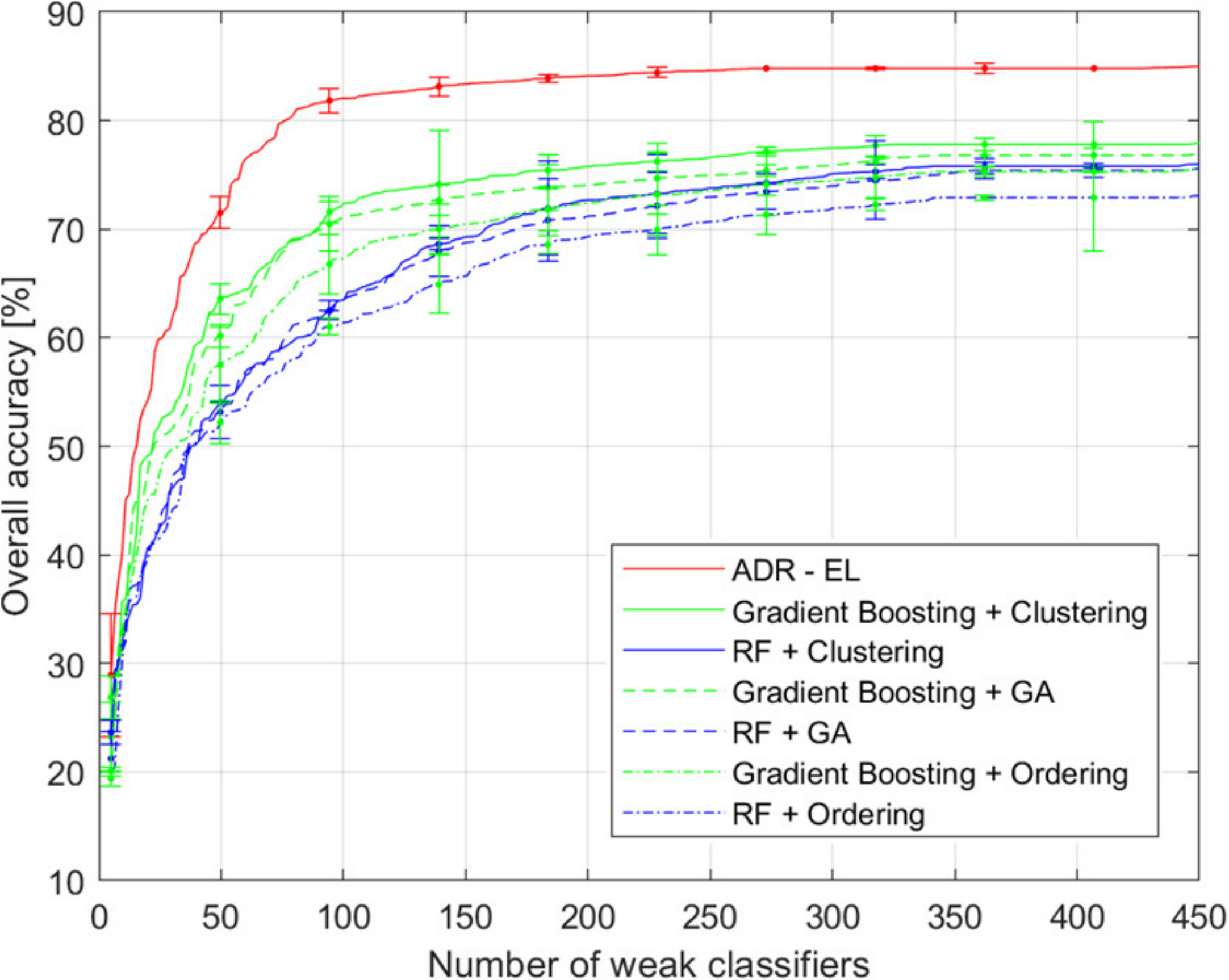} 
	\caption{Overall classification accuracy achieved over the multimodal remote sensing dataset in Section \ref{sec_exp_RS} by means of ensemble learning architectures at model level as a function of the number of weak classifiers that have been employed. The results obtained by means of the proposed architecture in Section \ref{sec_meth_RF}, Gradient Boosting, and Random Forest are displayed in red, green, and blue lines, respectively. The results that have been obtained when clustering, Genetic Algorithm (GA), and Ordering pruning techniques were used are shown in solid, dashed, and dash-dotted lines, respectively. Standard deviation measured over 100 experiments is displayed as error bars in the graph.}
	\label{fig_res_RF}
\end{figure}

\begin{figure}[htb]
	\centering
	\includegraphics[width=1\columnwidth]{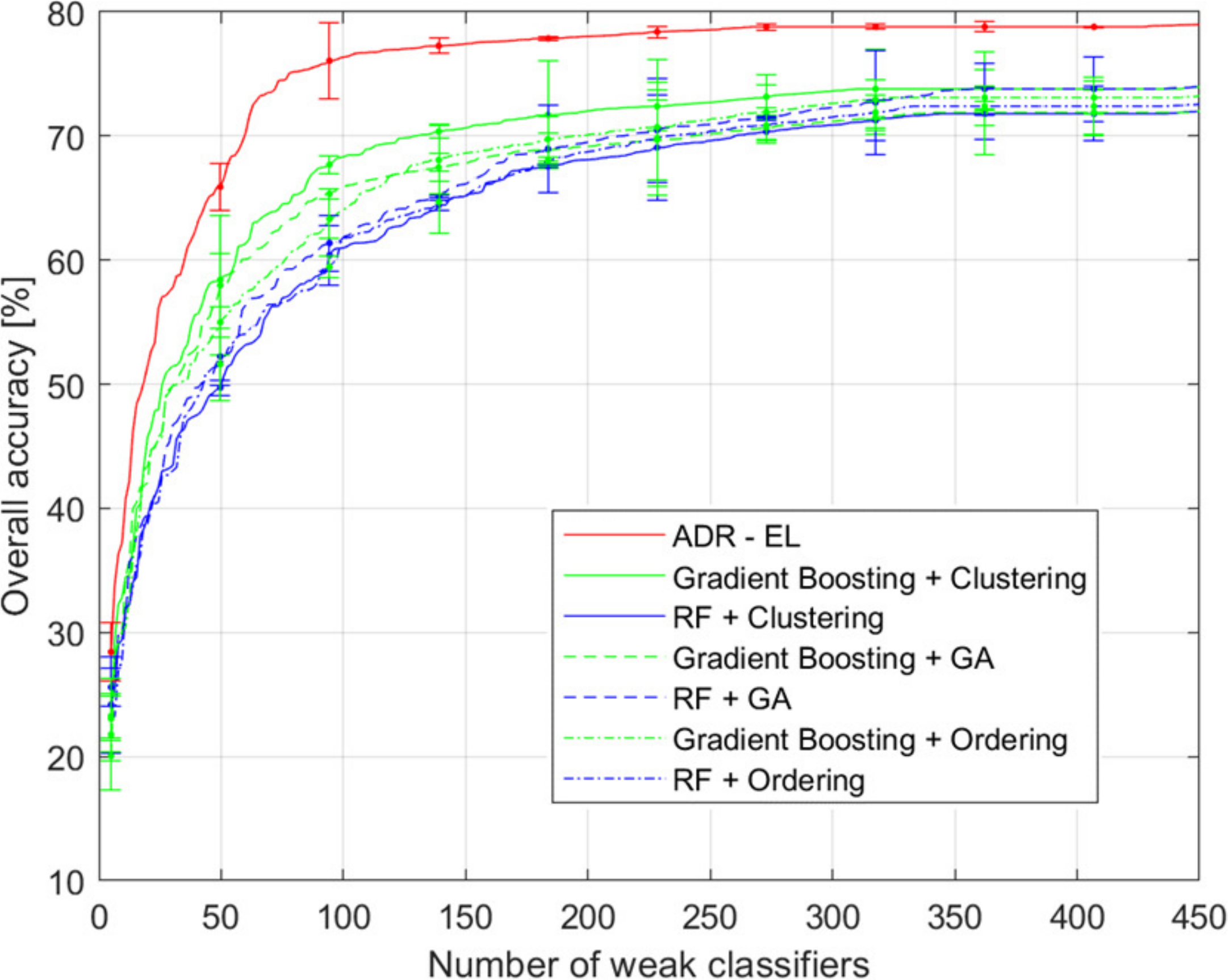} 
	\caption{Overall classification accuracy achieved over the multimodal brain-computer interface  dataset in Section \ref{sec_exp_BCI} by means of ensemble learning architectures at model level as a function of the number of weak classifiers that have been employed. The same notation as in Fig. \ref{fig_res_RF} applies here.}
	\label{fig_res_RF_BCI}
\end{figure}

\begin{figure}[htb]
	\centering
	\includegraphics[width=1\columnwidth]{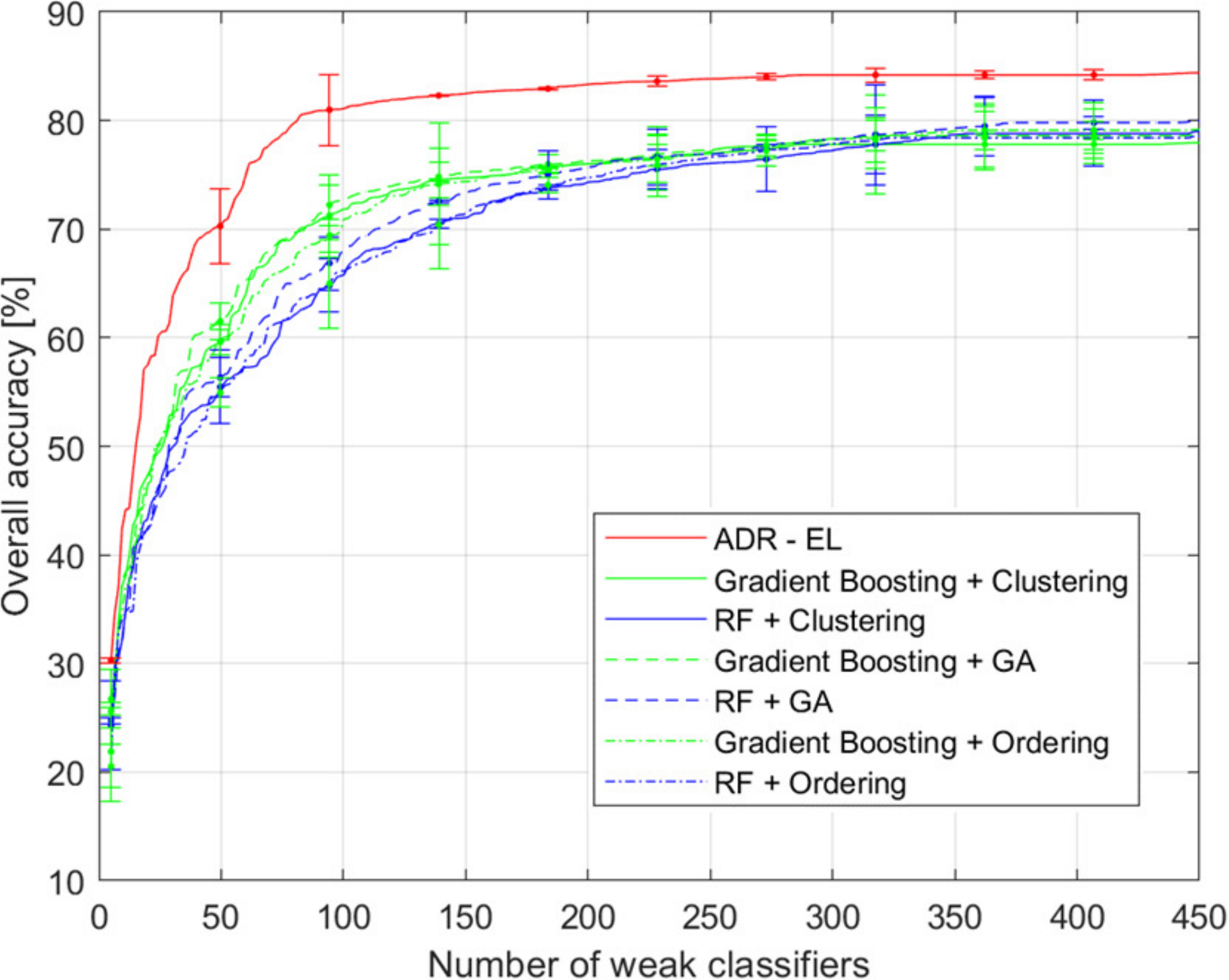} 
	\caption{Overall classification accuracy achieved over the multimodal photovoltaic energy dataset in Section \ref{sec_exp_PV} by means of ensemble learning architectures at model level as a function of the number of weak classifiers that have been employed. The same notation as in Fig. \ref{fig_res_RF} applies here.}
	\label{fig_res_RF_PV}
\end{figure}

Fig. \ref{fig_res_RF}, \ref{fig_res_RF_BCI} and \ref{fig_res_RF_PV} display the overall accuracy results we achieved as a function of the number of weak classifiers that have been employed on the multimodal remote sensing, brain-computer interface, and photovoltaic energy datasets, respectively. 
For fair comparison with the other  methods, the value on the x-axis for the proposed architecture identifies the total number of decision trees that have been implemented (i.e., $\Lambda M$ in Section \ref{sec_meth_RF}). 
The results obtained by means of the proposed architecture in Section \ref{sec_meth_RF} (ADR-EL), Gradient Boosting, and Random Forest are displayed in red, green, and blue lines, respectively. 
Moreover, pruning strategies have been applied with the gradient boosting and random forest methods: specifically, we considered clustering, genetic algorithm (GA), and ordering algorithms, and we display the corresponding results in solid, dashed, and dash-dotted lines, respectively. 
Finally, the standard deviation that has been measured over 100 experiments is displayed as error bars in the graphs.

It is possible to appreciate how the proposed architecture outperforms the state-of-the-art methods across the considered datasets.
It is interesting to note how the architecture introduced in Section \ref{sec_meth_RF} is able to achieve higher accuracy than gradient boosting and random forest for all the weak classifiers set-ups that have been considered. 
Moreover, it is worth noting that the proposed approach is able to converge (i.e., reach the maximum accuracy result) faster than the other methods. 
This point is particularly relevant for near real time application, when the efficiency of the system is highly stressed and addressed. 
Furthermore, it is possible to highlight how the proposed method does not show very high variability in the outcomes, especially when more than 100 weak classifiers (i.e., when the convergence of the method has already been achieved) are considered. 
This is a very important result, as it can be used to measure the precision of the proposed approach with respect to different set-ups. 
Moreover, taking a look to the average execution times obtained by the considered methods reported in Table \ref{tab_RF_ET}, it is possible to appreciate that the overall complexity of the proposed approach is fully comparable with the latency required by the state-of-the-art architectures, since the execution times over the considered datasets are very close.

\begin{table}[htb]
	\renewcommand{\arraystretch}{1.2}
	\begin{center}
		\caption{Average execution time [$sec$] for the model-level ensemble learning  experiments in Fig. \ref{fig_res_RF}, \ref{fig_res_RF_BCI}, and \ref{fig_res_RF_PV} [notation: gradient boosting (GB), random forest (RF), genetic algorithm (GA)]}
		\label{tab_RF_ET}
		\centering
		%\small
		\begin{tabular}{|c|c|c|c|}
			\hline
			\textbf{Method} & \textbf{RS} & \textbf{BCI} & \textbf{PV}  \\
			\hline
			\textbf{ADR - EL} & 169.4 $\pm$ 13.2& 175.8 $\pm$ 13.6 & 179.8 $\pm$14.2 \\
			\cline{1-4}
			\textbf{GB + Clustering} & 179.3 $\pm$ 13.6 & 172.4 $\pm$ 14.5	& 180.1 $\pm$ 16.3 \\
			\cline{1-4}
			\textbf{RF + Clustering} &174.1 $\pm$ 16.7& 170.2 $\pm$ 14&	171.7 $\pm$ 14.8				\\
			\cline{1-4}
			\textbf{GB + GA} &185.3 $\pm$ 17.4	& 180.1 $\pm$ 18.2&	182.3 $\pm$ 16.8
			\\
			\cline{1-4}
			\textbf{RF + GA} & 187 $\pm$ 18.4	& 185.6 $\pm$ 18.8	& 186.4 $\pm$ 17.3
			\\
			\cline{1-4}
			\textbf{GB + Ordering} & 174.8 $\pm$ 14.5 & 173.6 $\pm$ 12.4 & 176.7 $\pm$ 17.4 \\
			\cline{1-4}
			\textbf{RF + Ordering} & 176.9 $\pm$ 18.2 & 175.7 $\pm$ 14.9& 178.7 $\pm$ 15.4 \\
			\cline{1-4}	
			\hline
			
		\end{tabular}
	\end{center}
\end{table}

\subsubsection{Ensemble learning at data level}
\label{sec_exp_SVM}
We then analyzed the performance of the ensemble learning architecture developed at data level by considering the SVM scheme fed according to the guidelines summarized in Section \ref{sec_meth_SVM}. 
Specifically, for each supersample ${\cal S}_m$ we carried out a classification by means of an SVM algorithm that took into account only $\Omega_m$, i.e., the subset of features selected by the adaptive dimensionality reduction scheme proposed in Section \ref{sec_meth_ADR} for ${\cal S}_m$.
Each SVM exploits
radial basis functions with penalty parameter
of error term and 
 kernel coefficient automatically
selected via cross-validated grid-search optimization \cite{SVMpattern1}. 
Further, we ran 100 experiments by changing the training sets within the supersamples, in a classic 20-80 split between training and test sets. 

To validate and assess the actual classification accuracy performance obtained across the classes in the remote sensing Houston dataset, we compared the proposed architecture with schemes relying on dimensionality reduction algorithms representative of the approaches that are most commonly employed in technical literature:

\begin{itemize}
	\item Fisher score for attribute selection (FIS), as representative of ranking methods for dimensionality reduction \cite{FIS};
	\item principal component analysis (PCA), and decision boundary feature extraction (DBFE), as representative of feature extraction methods \cite{Theodoridis:2008:PRF:1457541, Xia16};
	\item forward attribute selection (FS), orthogonal branch and bound (OBB), and genetic algorithm (GA), representing searching strategies for dimensionality reduction \cite{Fauvel2015, Somol2000, Sivakumar2014};
	\item spectral clustering (SC), minimum spanning tree clustering (MST), dominant set (DS), and community detection (CD) as graph-based dimensionality reduction methods \cite{CD, specclust1, MST,DS}.
\end{itemize}

We compared these algorithms in terms of classification accuracy. In order to provide a fair comparison of
the different schemes, we implemented a common platform for
classification based on two algorithms, support vector machine
(SVM) and random forest (RF). %, widely used by the remote
% sensing community 
%(see, for instance, [54] and [55]). 
In this way, we can provide a shared platform for the
aforementioned schemes of dimensionality reduction so that consistent
evaluation of the performance can be conducted. %; on the other
%hand, it gives also the opportunity to quantitatively address
%the level of applicability of the image selection architectures
%to different frameworks of remote sensing image classification.
When RF is employed, the number of decision trees was
empirically set to 100 decision trees, and 
$\lfloor \sqrt{F} \rfloor ^2$ variables
are randomly drawn at each node of the trees, where F is the
number of features associated with each pixel that has been
identified by the aforementioned selection architectures \cite{randforPR4}. 
The SVM setup is analogous to the one we used for our proposed approach. 
%SVMs exploit radial basis functions [58], with parameters
%C (penalty parameter of error term) and γ (kernel coefficient) automatically selected via cross-validated grid-search
%optimization [18].

We also compared our approach with the ensemble SVM routine described in \cite{ELdata_pruning6}, where RBF is used as kernel function and optimal parameter selection is performed through cross-validation.  
Finally, we compared our approach with a deep learning-based scheme for spectral clustering and classification, SpectralNet \cite{specclust3}. 
In this case, in order to provide a fair and consistent comparison, we set the number of clusters the scheme aims to identify to $M$, i.e., the number of supersamples identified by the proposed adaptive dimensionality reduction scheme. 
Moreover, we set the affinity between labeled points according to their true labels, so to use SpectralNet as classifier. 

The classification performance results we have obtained by using the aforesaid methods and the proposed approach on the datasets described in Sections 
%\ref{sec_exp_RS}, \ref{sec_exp_BCI}, and \ref{sec_exp_PV} 
\ref{sec_exp_RS}-\ref{sec_exp_PV}
are summarized in Table \ref{tab_houston}, where we report the average classification accuracy (as a percentage) and the corresponding standard deviation measured for every architecture.
It is hence possible to appreciate that the architecture in Section \ref{sec_meth_SVM} is able to solidly outperform the other methods. 
Moreover, it is worth noting how the proposed scheme shows a value of standard deviation typically smaller than the methods that have been tested. 
This emphasizes the high stability of the proposed method, for which it is possible to expect the proposed approach to appropriately address the variability of the data without suffering strong degradation in terms of classification accuracy. 
On the other hand, these outstanding accuracy results come at a cost in terms of 
the average execution time (reported in seconds in Table \ref{tab_houston}) registered for the experiments we conducted, since some state-of-the-art architectures require lower latency to analyze the considered datasets. 
In fact, it is possible to appreciate that methods relying on graph-based dimensionality reduction (i.e., SC, MST, DS, CD, SpectralNet, EnsembleSVM and ADR-SVM) typically show less latency than the other algorithms. 
It is also true that some of the methods proposed in technical literature can take advantage of high performance computing systems (e.g., SpectralNet) and optimized programming based on dedicated libraries (e.g., CD).  
Nevertheless, since the proposed scheme takes advantage of matricial operations, 
the proposed scheme could be adequate to be implemented on parallel computing platforms (out of the scope of this work). 
Thus,  it is possible to state that the ET results could be meliorated in this case, making the proposed approach a contender to the state-of-the-art methods in terms of latency too.

\begin{table*}[htb]
	\renewcommand{\arraystretch}{1.5}
	\begin{center}
		\caption{Overall classification accuracy [$\%$] and computational analysis [sec] obtained over the considered datasets: average ($\mu$), standard deviation ($\sigma$) and average execution time (ET) [$sec$] over 100 experiments. The best results are shown in blue color.}
		\label{tab_houston}
		\centering
		%\small
		\begin{tabular}{|c|c|c|c|c|c|c|c|c|c|c|}
			\hline
			\multicolumn{2}{|c|}{} & 
			\multicolumn{3}{c|}{\textbf{RS}} &
			\multicolumn{3}{c|}{\textbf{BCI}} &
			\multicolumn{3}{c|}{\textbf{PV}}
			\\
			\multicolumn{2}{|c|}{\textbf{Method}}& \textbf{$\mu$} & \textbf{$\sigma$}  & ET &
			\textbf{$\mu$} & \textbf{$\sigma$}  & ET &
			\textbf{$\mu$} & \textbf{$\sigma$}  & ET \\
			\hline
			\hline
			
			\multirow{2}{*}{\textbf{PCA}} & \textbf{RF} & 79.6 & 3.2 & 188 & 70&2 & 150& 75 & 2.3& 187\\
			\cline{2-11}
			& \textbf{SVM} & 83.2 & 4.1 & 98 & 71.1& 2.9& 115 &76.2 &2.8 & 166\\
			\cline{1-11}
			
			\multirow{2}{*}{\textbf{DBFE}} & \textbf{RF} & 78.9 & 2.7& 196 & 72&2.4 & 137&74.2 &2.5 & 193\\
			\cline{2-11}
			& \textbf{SVM} & 79.4 & 4& 99 &71.8 & 1.9 & 114 &73.4 &3.5 & 182\\
			\cline{1-11}
			
			\multirow{2}{*}{\textbf{FIS}} & \textbf{RF} & 80.3 & 3.3 & 189 &70.1 &2.4 & 129&72.3 &2.6 & 182\\
			\cline{2-11}
			& \textbf{SVM} & 80.6 & 3.8&103 &70.4 &2.5 & 112 &71.7 & 2.6& 165\\
			\cline{1-11}
			
			\multirow{2}{*}{\textbf{FS}} & \textbf{RF} & 77.5 & 2.9 & 190 & 70.2&2.5 & 177&73.8 &2.9 & 188\\
			\cline{2-11}
			& \textbf{SVM} & 77.2 & 2.7 & 101 &70.4 &2.6 & 124 &74.5 &2.1 & 161\\
			\cline{1-11}
			
			\multirow{2}{*}{\textbf{OBB}} & \textbf{RF} & 78.5 & 3.2 & 209&71.3 &2.2 & 166 &73.1 &2.4 &199\\
			\cline{2-11}
			& \textbf{SVM} & 77.2 & 3.5 & 116 &72.7 &2.3 & 136 &72.8 &2.4 &173\\
			\cline{1-11}
			
			\multirow{2}{*}{\textbf{GA}} & \textbf{RF} & 80.2 & 2.4 & 200 & 73&2.2 & 160 & 76.5&2.7 & 188\\
			\cline{2-11}
			& \textbf{SVM} & 78.6 & 2.3 & 111 &71.9 &2.9 & 123 &76.6 &2.6 &160\\
			\cline{1-11}
			
			\multirow{2}{*}{\textbf{SC}} & \textbf{RF} & 86.2 & 1.6 & 61 & 74.1 &2.3 & 55& 77.9&2.7 & 62\\
			\cline{2-11}
			& \textbf{SVM} & 87.5 & 1.7 & 59 & 75.2& 2.4& 54&78 &2.3 & 58\\
			\cline{1-11}
			
			\multirow{2}{*}{\textbf{MST}} & \textbf{RF} & 83.6 & 2.4 & 54 & 73.3& 2.5& 44 &76.7 &2.2 &53\\
			\cline{2-11}
			& \textbf{SVM} & 85.7 & 2.3  & 43 & 73.9 &2.4 & 39&77.5 &2.4 & 48\\
			\cline{1-11}
			
			\multirow{2}{*}{\textbf{DS}} & \textbf{RF} & 79.8 & 2.5 & 45 &72.7 &2.6 & 36&76.4 &2.7 &55\\
			\cline{2-11}
			& \textbf{SVM} & 77.5 & 2.1 & \color{blue} \textbf{31} \color{black} &72.3 &2.4 & \color{blue}\textbf{28} \color{black} &75.8 &2.7 &49\\
			\cline{1-11}
			
			\multirow{2}{*}{\textbf{CD}} & \textbf{RF} & 82.8 & 1.8 &52 &71.7 &2.6 & 46&75.6 &2.3 &56\\
			\cline{2-11}
			& \textbf{SVM} & 85.3 & 2.1 &47 & 71.2& 2.4& 41&74.7 &2.8 &48\\
			\cline{1-11}
			
			\multicolumn{2}{|c|}{\textbf{SpectralNet}} & 88.1 & 1.6 &46 & 77.7& \color{blue}\textbf{1.5}\color{black}& 32& 81.8 &1.7 &\color{blue}\textbf{42} \color{black}\\
			\hline
			
			\multicolumn{2}{|c|}{\textbf{EnsembleSVM}} & 80.3 & 3.2 & 55& 75.9&2 & 35& 78.8&2.4 & 59\\
			\hline
			
			\multicolumn{2}{|c|}{\textbf{ADR - SVM}} & \color{blue} \textbf{91.6} \color{black} & \color{blue} \textbf{1.4} \color{black} & 64& \color{blue} \textbf{80.2} \color{black}& 1.6 & 59& \color{blue} \textbf{84.4} \color{black} & \color{blue} \textbf{1.2} \color{black} & 69 \\
			\hline

			\hline
			%			\multirow{3}{*}{LMM} & 0.6 & 7 & 7.7 & 7.7 & 5 \\
			%			\cline{2-6}
			%			& 0.75 & 5.9 & 6.3 & 6.2 & 4.8 \\
			%			\cline{2-6}
			%			& 0.9 & 4.8 & 5.2 & 5 & 4.6 \\
			%			\cline{1-6}
			%			
			%			\multirow{3}{*}{LQM} & 0.6 & 19.9 & 7.6 & 7.5 & 5.2 \\
			%			\cline{2-6}
			%			& 0.75 & 16.9 & 6.8 & 6.8 & 4.8 \\
			%			\cline{2-6}
			%			& 0.9 & 15.8 & 5.7 & 5.6 & 4.7 \\
			%			\cline{1-6}
			%			
			%			\multirow{3}{*}{3LMM} & 0.6 & 24.04 & 9.6 & 7.6 & 5.1 \\
			%			\cline{2-6}
			%			& 0.75 & 22.7 & 8 & 6.2 & 4.8 \\
			%			\cline{2-6}
			%			& 0.9 & 20.8 & 6.9 & 4.8 & 4.6 \\
			%			\cline{1-6}
			%			
			%			\multirow{3}{*}{3CMM} & 0.6 & 34.8 & 17.5 & 17.6 & 4.4 \\
			%			\cline{2-6}
			%			& 0.75 & 32.6 & 14.8 & 14.7 & 4.4 \\
			%			\cline{2-6}
			%			& 0.9 & 29.8 & 12 & 11.9 & 4.2 \\
			%			\cline{1-6}
			%			
			%			\hline
			%%%
			
		\end{tabular}
	\end{center}
\end{table*}

\begin{table}[htb]
	\renewcommand{\arraystretch}{1.2}
	\begin{center}
		\caption{Average execution time [$sec$] for the transductive transfer learning experiments in Fig. \ref{fig_res_TTL}, \ref{fig_res_TTL_BCI}, and \ref{fig_res_TTL_PV}.}
		\label{tab_TTL_ET}
		\centering
		%\small
		\begin{tabular}{|c|c|c|c|}
			\hline
			\textbf{Method} & \textbf{RS} & \textbf{BCI} & \textbf{PV}  \\
			\hline
			\textbf{ADR - TTL} & 80 $\pm$ 2.3& 77.9 $\pm$ 3.1 & 81.5 $\pm$3.8 \\
			\cline{1-4}
			\textbf{ERM} &123 $\pm$ 11.9 &	111.2 $\pm$ 4.6	& 129.1 $\pm$ 10.3 \\
			\cline{1-4}
			\textbf{KMM} &164 $\pm$ 13.3& 126 $\pm$ 5.8&	170.3 $\pm$ 8.9				\\
			\cline{1-4}
			\textbf{KLIEP} &177 $\pm$ 12.7	& 168 $\pm$ 5.2&	179.8 $\pm$ 10.4
			\\
			\cline{1-4}
			\textbf{DANN} & 81 $\pm$ 4.4	& 78.8 $\pm$ 6	& 80.2 $\pm$ 5.6
			\\
			\cline{1-4}
			\textbf{MDDA} & 86 $\pm$ 2.5 & 80.6 $\pm$ 3.7 & 85.8 $\pm$ 4 \\
			\cline{1-4}
			\textbf{DDAN} & 70.2 $\pm$ 4.8 & 65.7 $\pm$ 3.8& 71.2 $\pm$ 5.1 \\
			\cline{1-4}	
			\textbf{DRCN} & 71.3 $\pm$ 3.2	& 67.9 $\pm$ 6.9	& 72.3 $\pm$ 3.7\\
			\cline{1-4}	
			\hline
			
		\end{tabular}
	\end{center}
\end{table}

\subsubsection{Transductive transfer learning}
\label{sec_exp_TTL}
Finally, we tested the classification accuracy performance of the proposed scheme for transductive tranfer learning introduced in Section \ref{sec_meth_TTL}. 
We compared the proposed approach with three of the major transductive transfer learning algorithms that we introduced in Section \ref{sec_methback}: empirical risk
minimization (ERM) \cite{TTL2},
kernel-mean matching (KMM) algorithm \cite{TTL3},
Kullback-Leibler importance estimation procedure (KLIEP) \cite{TTL4}, domain adversarial neural network (DANN) \cite{TTL5}, manifold dynamic domain adaptation (MDDA) \cite{TTL8}, dynamic domain adaptation network (DDAN) \cite{TTL8}, and deep residual correction network (DRCN) \cite{TTL9}. 
Specifically, we ran several tests by changing the size of the source domain %(in terms of fraction of samples, which can be written as $|X^{\text{S}}|/P$, according to the notation used in Section \ref{sec_meth_TTL}). 
Moreover, the samples in $\underline{\underline{X}}^{\cal S}$ were randomly selected 100 times, uniformly across the classes. 

Fig. \ref{fig_res_TTL}, \ref{fig_res_TTL_BCI} and \ref{fig_res_TTL_PV} show the overall accuracy results we achieved for the aforementioned architectures as a function of the source domain size. 
The average overall accuracy obtained by using ERM, KMM, KLIEP, DANN, MDDA, DDAN, and DRCN algorithms is shown colors across the parula colormap. 
The outcomes of the scheme in Section \ref{sec_meth_TTL} are shown in red. 
Error bars display the standard deviation that has been measured across the 100 experiments we carried out for the different values of source domain size. 
By taking a look to the trends that are shown in the figure, it is possible to appreciate that the proposed approach is able to provide higher accuracy in the classification of the considered dataset.
Moreover, the proposed architecture is apparently able to achieve classification performance close to the maximum with a pretty low amount of labeled samples in the source domain (approximately 30 $\%$ of the total dataset size), while the other methods do not show saturating trends until the source domain gets very large. 
%very high values of $|X^{\text{S}}|/P$. 
The variability of the classification accuracy results in the proposed architecture decreases as the source domain size increases to values that are in any case smaller than those registered for the methods in technical literature we took into account. 
Finally, Table \ref{tab_TTL_ET} summarizes the average execution time result achieved by the aforementioned architectures. 
Apparently, the proposed scheme shows comparable performance with respect to the state-of-the-art algorithms across the three considered datasets. 

\subsubsection{Impact of adaptive dimensionality reduction}
These results, together with the main characteristics and properties of the competing state-of-the-art methods, emphasize how the selection of relevant features in multimodal datasets performed in an adaptive fashion (as explained in Section \ref{sec_meth_ADR}) can provide a strong impact in the characterization of the considered datasets, so to solidly enhance the understanding of the samples to be investigated. 
To further support this point, we have conducted several tests to assess the actual impact of the proposed approach for adaptive dimensionality reduction over the datasets and experiments that have been previously reported. 
In particular, we focused our attention to evaluate the enhancement provided by ADR method over the use of spectral clustering relying on a weight matrix computed according to kernels based on the Gaussian distribution and the mutual information function. 
In other terms, we compared the performance obtained by the methods in Sections \ref{sec_meth_RF}, \ref{sec_meth_SVM} and \ref{sec_meth_TTL} using ADR with respect to the same algorithms when feature selection is performed by spectral clustering based on Laplacian matrices resulting by the employment of the weights in (\ref{eq_GKMI1}) and (\ref{eq_GKMI2}), respectively. 

\begin{figure}[htb]
	\centering
	\includegraphics[width=1\columnwidth]{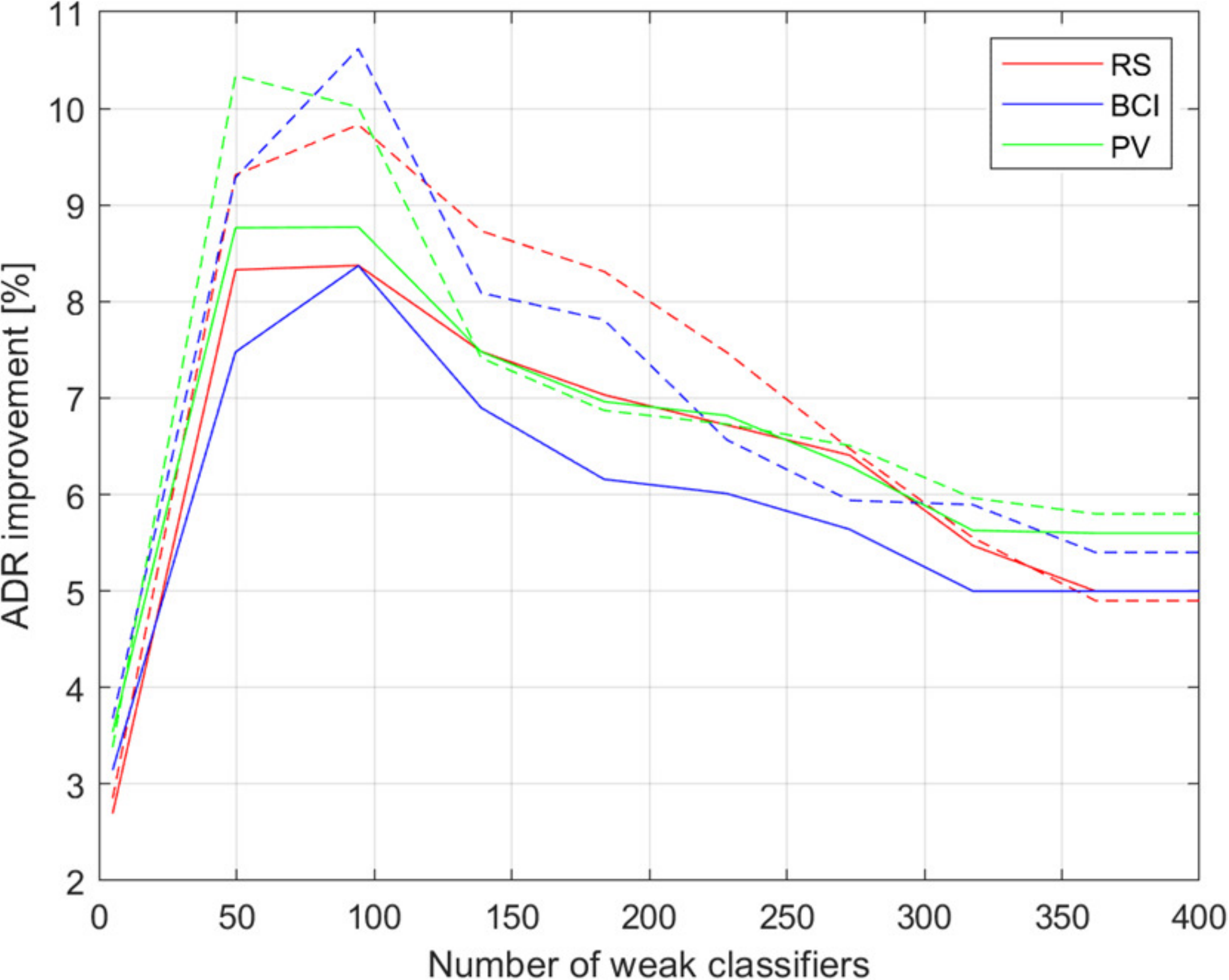} 
	\caption{Overall accuracy improvement provided by the use of adaptive dimensionality reduction when the ensemble learning architecture in Section \ref{sec_meth_RF} is employed with respect to dimensionality reduction conducted by using spectral clustering based on Gaussian kernel (solid lines) and mutual information kernel (dashed lines). The results obtained when considering multimodal remote sensing (RS), brain computer interface (BCI), photovoltaic energy (PV) datasets are reported in red, blue, and green color, respectively.}
	\label{fig_res_RF_GK_MI}
\end{figure}

In this context, we focused our attention on the ensemble learning architecture at model level investigated in Section \ref{sec_exp_RF}. 
We computed the improvement achieved by using the ADR algorithm over the employment of spectral clustering based on Gaussian kernel and mutual information. 
The results we obtained when considering the datasets in Section \ref{sec_exp_RS}-\ref{sec_exp_PV} are shown in Fig. \ref{fig_res_RF_GK_MI} in red, blue and green color, respectively. Further, the outcomes obtained when considering spectral clustering based on Gaussian kernel and mutual information weight matrices are reported in solid and dashed lines, respectively. 
It is then possible to appreciate that the proposed architecture for adaptive dimensionality reduction can actually provide a solid enhancement of the accuracy performance. 
The improvement gained by ADR is stronger when fewer weak classifiers are considered. 
Also, it is interesting to note that the larger improvement is typically registered with respect to spectral clustering based on mutual information. This result is particularly relevant because the weight matrix relying on the metric in (\ref{eq_GKMI2}) is computed across each modality for each sample in the dataset. 
Nevertheless, the use of spectral clustering based on Gaussian kernel does not deliver much stronger performance, since the gain provided by ADR is still substantial for all the values of weak classifiers  and for all the datasets considered. 

The aforementioned results show how combining Gaussian kernel and mutual information in one single metric to achieve adaptive dimensionality reduction is a key factor to boost the accuracy performance of the ensemble learning architectures. 
This statement is further confirmed by taking a look to Table \ref{tab_SVM_GK_MI}, where the overall improvement provided by ADR over Gaussian kernel- and mutual information-based spectral clustering when considering the methods in Section \ref{sec_exp_SVM} is reported. 
Once again, the enhancement provided by integrating global (mutual information) and local (Gaussian kernel) scale metrics to run the dimensionality reduction in an adaptive fashion represents a winning choice with respect to the separate use of these distances to drive spectral clustering for feature selection. 

\begin{figure}[htb]
	\centering
	\includegraphics[width=1\columnwidth]{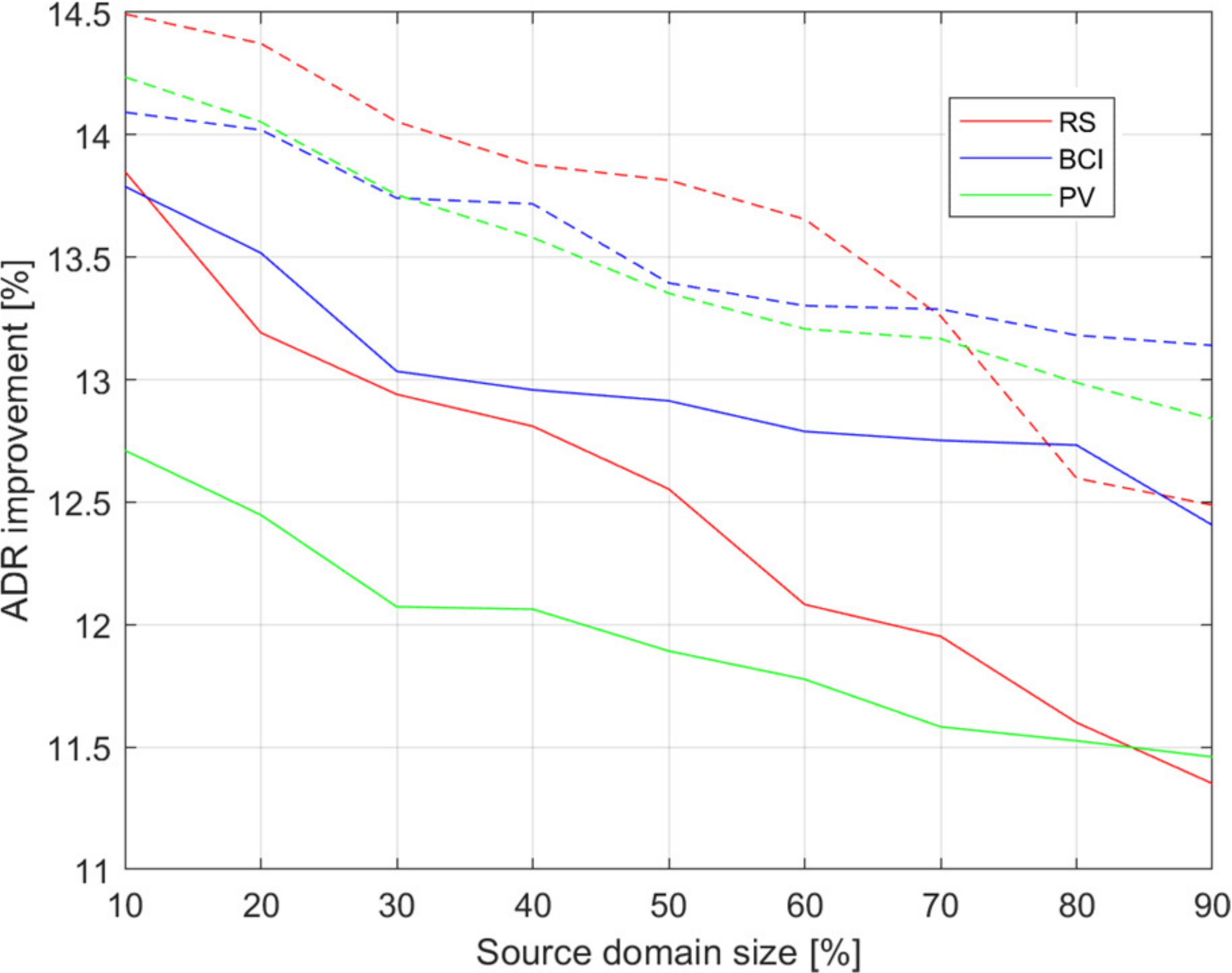} 
	\caption{Overall accuracy improvement provided by the use of adaptive dimensionality reduction when the ensemble learning architecture in Section \ref{sec_meth_TTL} is employed with respect to dimensionality reduction conducted by using spectral clustering based on Gaussian kernel (solid lines) and mutual information kernel (dashed lines). The results obtained when considering multimodal remote sensing (RS), brain computer interface (BCI), photovoltaic energy (PV) datasets are reported in red, blue, and green color, respectively.}
	\label{fig_res_TTL_GK_MI}
\end{figure}

Analogously, when considering the situation described in Section \ref{sec_exp_TTL}, the improvement provided by using ADR still appears solid. 
This statement is supported by the results shown in Fig. \ref{fig_res_TTL_GK_MI}, where the same notation as in Fig. \ref{fig_res_RF_GK_MI} applies. 
As such, it is possible to appreciate that using ADR substantially enhances the accuracy performance of the transfer learning structure with respect to the case for which dimensionality reduction is conducted by means of Gaussian kernel- and mutual information-based spectral clustering. 
In this case, it is interesting to note that the largest improvement provided by ADR is obtained when the size of the source domain is small with respect to the overall dataset size. 
This result is particularly important, since it further proves the ability of the proposed architecture to provide high transfer learning performance also when the population of the source and target domains are very skewed. 
On the other hand, the separate use of the metrics in (\ref{eq_GKMI1}) and (\ref{eq_GKMI2}) does not provide any enhancement also in the transfer learning application. 
Thus, integrating the Gaussian kernel and mutual information metrics seems to be the best design choice we can take in order to address the main issues of multimodal data analysis in ensemble learning and transfer learning.

\begin{figure}[htb]
	\centering
	\includegraphics[width=1\columnwidth]{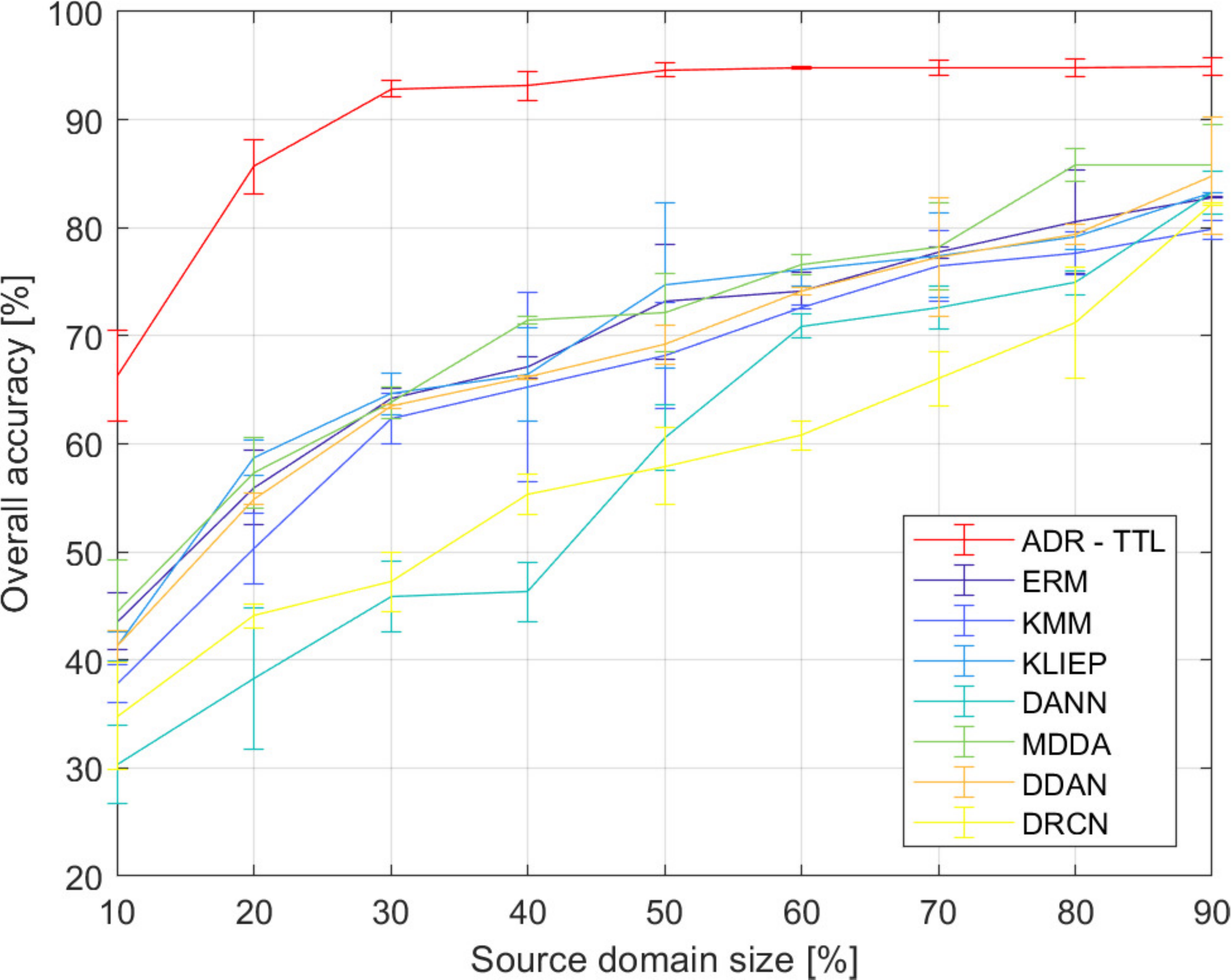} 
	\caption{Overall classification accuracy achieved over the multimodal remote sensing dataset in Section \ref{sec_exp_RS} by means of transductive tranfer learning architectures as a function of the size of the source domain. The results obtained by means of the proposed architecture in Section \ref{sec_meth_TTL} are displayed in red. The results obtained by means of ERM \cite{TTL2}, KMM \cite{TTL3}, KLIEP \cite{TTL4}, DANN \cite{TTL5}, MDDA, DDAN \cite{TTL8}, and DRCN \cite{TTL9} methods are shown in colors across the parula colormap. Standard deviation measured over 100 experiments is displayed as error bars in the graph.}
	\label{fig_res_TTL}
\end{figure}

\begin{figure}[htb]
	\centering
	\includegraphics[width=1\columnwidth]{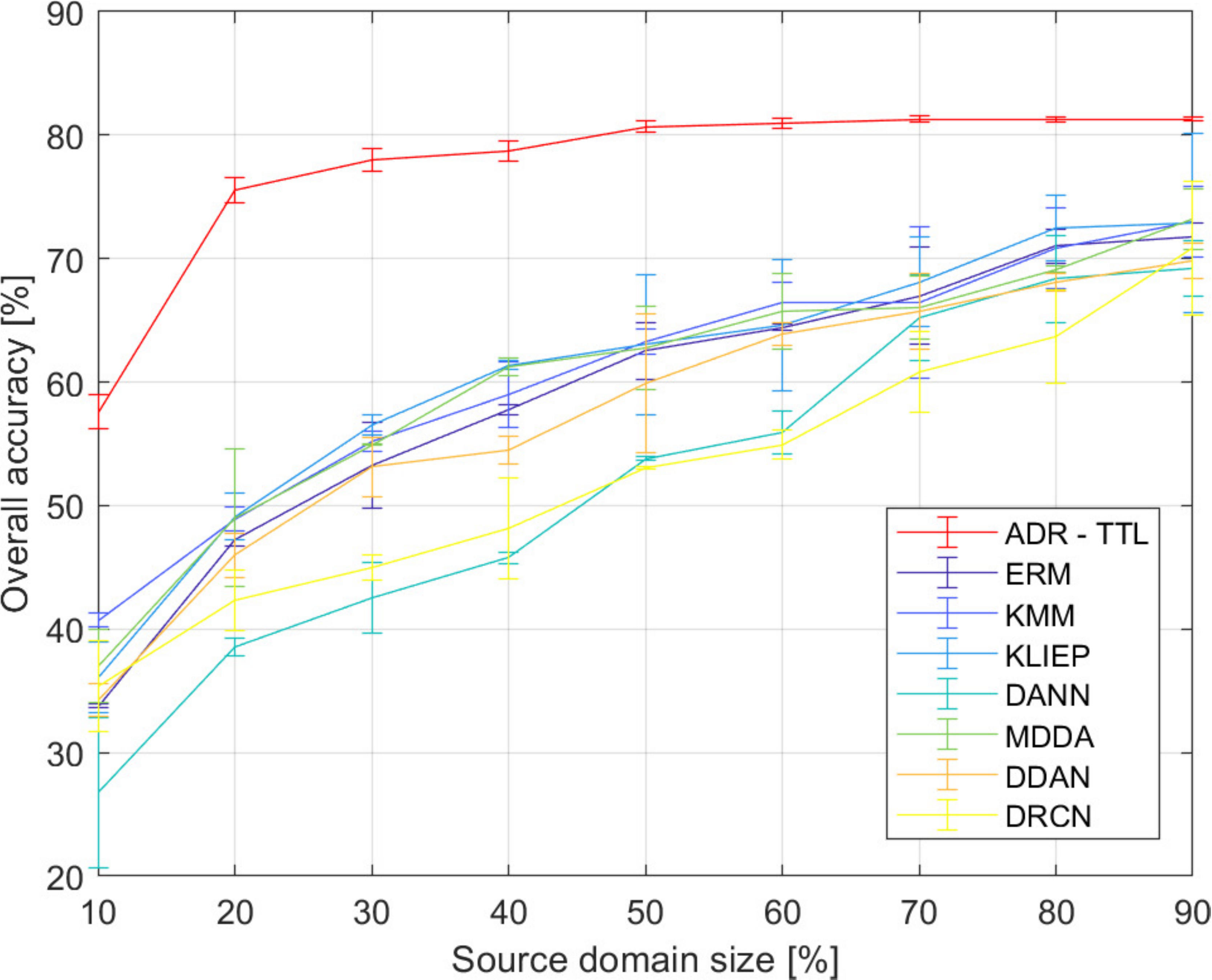} 
	\caption{Overall classification accuracy achieved over the multimodal brain-computer interface dataset in Section \ref{sec_exp_BCI} by means of transductive tranfer learning architectures as a function of the size of the source domain. The same notation as in Fig. \ref{fig_res_TTL} applies here.}
	\label{fig_res_TTL_BCI}
\end{figure}

\begin{figure}[htb]
	\centering
	\includegraphics[width=1\columnwidth]{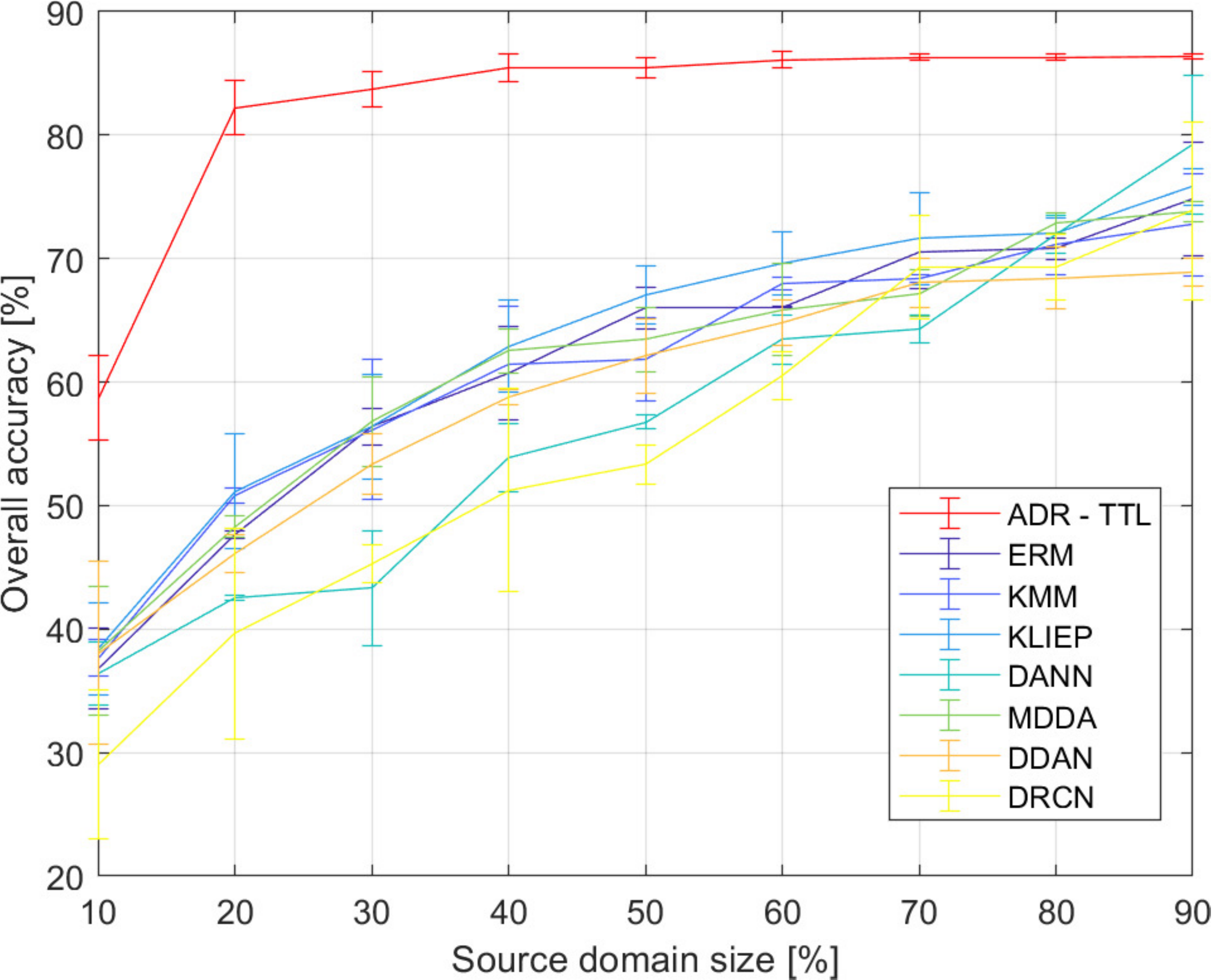} 
	\caption{Overall classification accuracy achieved over the multimodal photovoltaic energy dataset in Section \ref{sec_exp_PV} by means of transductive tranfer learning architectures as a function of the size of the source domain. The same notation as in Fig. \ref{fig_res_TTL} applies here.}
	\label{fig_res_TTL_PV}
\end{figure}

\begin{figure}[htb]
	\centering
	\includegraphics[width=1\columnwidth]{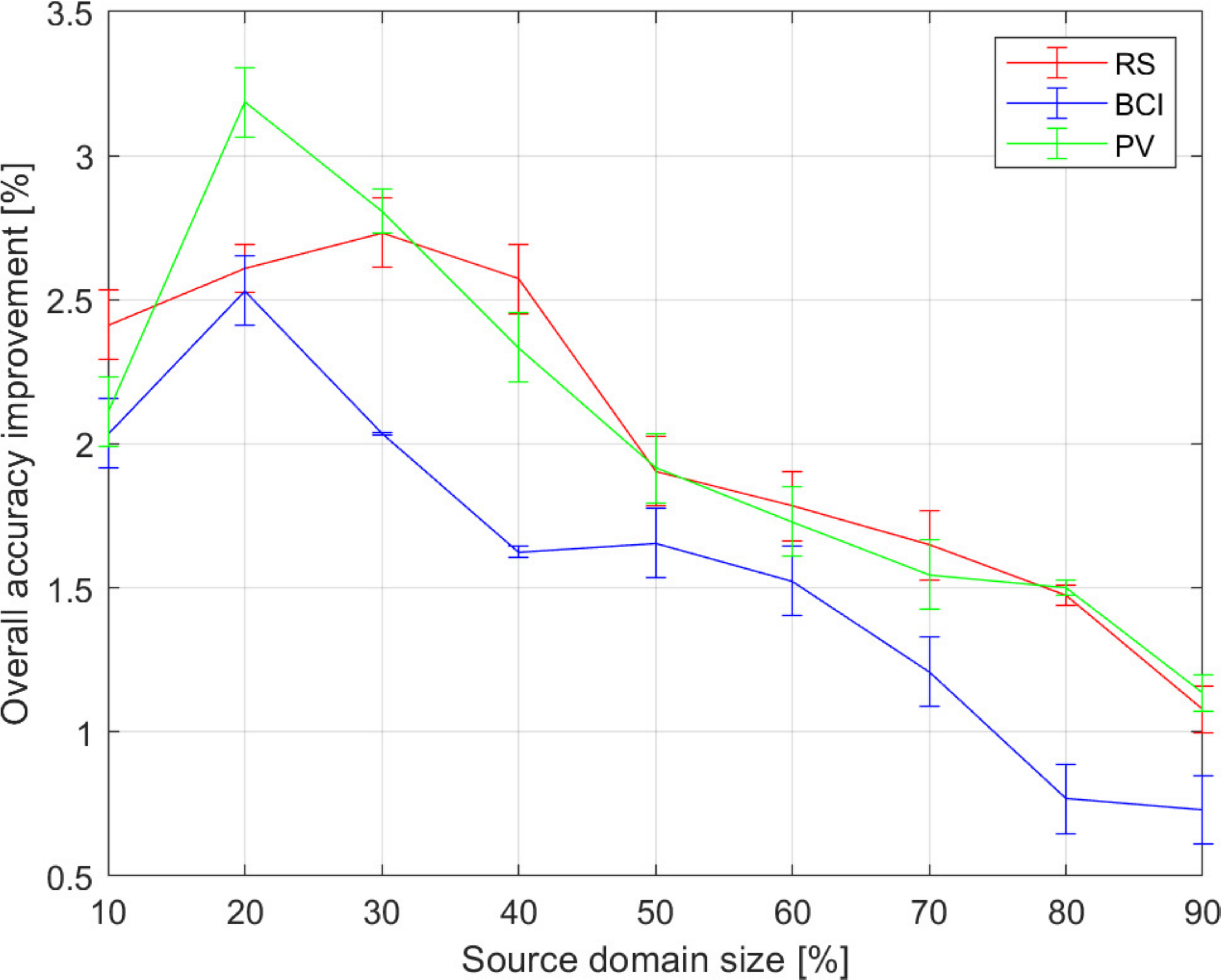} 
	\caption{Overall accuracy improvement of the KLIEP architecture \cite{TTL4} performance achieved by applying adaptive dimensionality reduction as a function of the size of the source domain over the datasets in Section \ref{sec_exp_data}. The results obtained when considering multimodal remote sensing (RS), brain computer interface (BCI), photovoltaic energy (PV) datasets are reported in red, blue, and green color, respectively.}
	\label{fig_res_TTL_KLIEP}
\end{figure}

Finally, we have carried out an additional analysis by slightly modifying the KLIEP architecture \cite{TTL4} to take into account the proposed adaptive dimensionality reduction scheme. 
In particular, we modified the set of model candidates (originally based on classic Gaussian kernels) so to take into account only the features that have been selected by the ADR procedure introduced in Section \ref{sec_meth_ADR}. 
The improvement in overall accuracy obtained over the datasets in Section \ref{sec_exp_data} is reported in Fig. \ref{fig_res_TTL_KLIEP}. 
Specifically, the results obtained when considering the multimodal remote sensing, brain-computer interface, and photovoltaic energy data are displayed in red, blue and green color, respectively. 
It is possible to appreciate that this modification provides larger enhancement of the accuracy performance when the size of the source domain is small. 
This result is particularly interesting, as it apparently links the adaptive dimensionality reduction algorithm to the rapid convergence showed also in Fig. \ref{fig_res_TTL}, \ref{fig_res_TTL_BCI} and \ref{fig_res_TTL_PV}. 
This effect is hence worth to be further investigated in future works. 
Moreover, it is worth noting that ADR could be integrated in other existing methods (e.g., ERM, KMM) as a preprocessing operation. 
Therefore, future works will be dedicated to investigate this other approach as well.

\begin{table}[htb]
	\renewcommand{\arraystretch}{1.2}
	\begin{center}
		\caption{Overall accuracy improvement [$\%$] provided by the use of adaptive dimensionality reduction when the ensemble learning architecture in Section \ref{sec_meth_SVM} is employed with respect to dimensionality reduction conducted by using spectral clustering based on Gaussian kernel (SC - GK) and mutual information kernel (SC - MI)}
		\label{tab_SVM_GK_MI}
		\centering
		%\small
		\begin{tabular}{|c|c|c|c|}
			\hline
			\textbf{Method} & \textbf{RS} & \textbf{BCI} & \textbf{PV}  \\
			\hline
			\textbf{SC - GK} & 4.1 & 5 & 6.4 \\
			\cline{1-4}
			\textbf{SC - MI} &4.8 &	7.1	& 7.7 \\
			\cline{1-4}
			
		\end{tabular}
	\end{center}
\end{table}

 \section{Conclusion}
 \label{secconcl}
 
 Multimodal data analysis has recently become a hot topic in the scientific community, especially thanks to its ability to retrieve higher level of details on several social, financial, environmental phenomena 
 This result is achieved by taking advantage of the diversity of the sources of information (modalities), 
 and integrating the information obtained by the multiple records to build knowledge for different research fields and applications. 
 To this aim, ensemble learning and transfer learning strategies can play a key role.
 In this paper, we report a novel approach to enhance the ability of these techniques by investigating the most relevant features in the considered datasets, hence improving the traditional approaches by addressing the reliability of the considered records.
 The main contributions of this work are:
 \begin{itemize}
 	\item the introduction of a method for adaptive dimensionality reduction based on a double graph Laplacian investigation;
 	\item  the improvement of ensemble learning and transfer learning schemes by a better characterization of analysis models and attributes to be used to describe the interactions among the samples;
 	\item the increase of robustness of multimodal data analysis by investigation of significant features, so that the data analysis schemes can reach convergence in fewer steps and with low variability and sensitivity to initial conditions.
 \end{itemize}

The experimental results achieved on three different application scenarios show the validity of our approach. 
Future works will be dedicated to further investigate the ability of the proposed approach to address several challenges in modern data analysis, such as limited amount of data for model training, quantification of uncertainty and reliability in domain adaptation and information propagation, and using graph-based processing to enhance the semantic interpretation of the considered datasets. 

 %as integrating the available multiple resources can help to characterize the 

\section{Acknowledgements}

This work is funded in part by Centre for Integrated Remote
Sensing and Forecasting for Arctic Operations (CIRFA) and
the Research Council of Norway (RCN Grant no. 237906),
 the Automatic Multisensor remote sensing for Sea Ice
Characterization (AMUSIC) Framsenteret ”Polhavet” flagship project 2020, the Isaac Newton Trust, and Newnham College, Cambridge, UK. The authors thank Theo Damoulas (University of Warwick and the Alan Turing Insitute, UK) for useful conversations. 

\bibliographystyle{unsrt}\scriptsize	
\bibliography{JPODrefs,BIGDATArefs}

\end{document}